\providecommand\FormatOverride{mlsys}              \providecommand\LogosOverride{camlsys-cambridge-flower} \providecommand\TitleBgOverride{camlsys}             \documentclass{paper}

\input{preamble/packages}
\input{preamble/macros}
\input{preamble/adapter}
\input{preamble/logos}

\paperTitle{\method: Breaking the Full-Replica Barrier with a Federation of MoEs}
\paperTitleRunning{\method: Breaking the Full-Replica Barrier with a Federation of MoEs}
\paperKeywords{machine learning systems, mixture of experts, federated learning, large language models}

\addAffiliation{cam}{Cambridge ML Systems Lab (CaMLSys), University of Cambridge}
\addAffiliation{flower}{Flower Labs}

\addAuthor{Lorenzo Sani}{cam,flower}
\addAuthor{Zeyu Cao}{cam}
\addAuthor{Meghdad Kurmanji}{cam}
\addAuthor{Alex Iacob}{cam,flower}
\addAuthor{Andrej Jovanović}{cam,flower}
\addAuthor{Yan Gao}{cam,flower}
\addAuthor{Wanru Zhao}{cam}
\addAuthor{Nicholas D.\ Lane}{cam,flower}

\correspondingAuthor{Lorenzo Sani}{ls985@cam.ac.uk}

\begin{document}

\makepapertitle

\section{Introduction}\label{sec:intro}

\begin{figure}[htb]
    \centering
    \includegraphics[width=\linewidth]{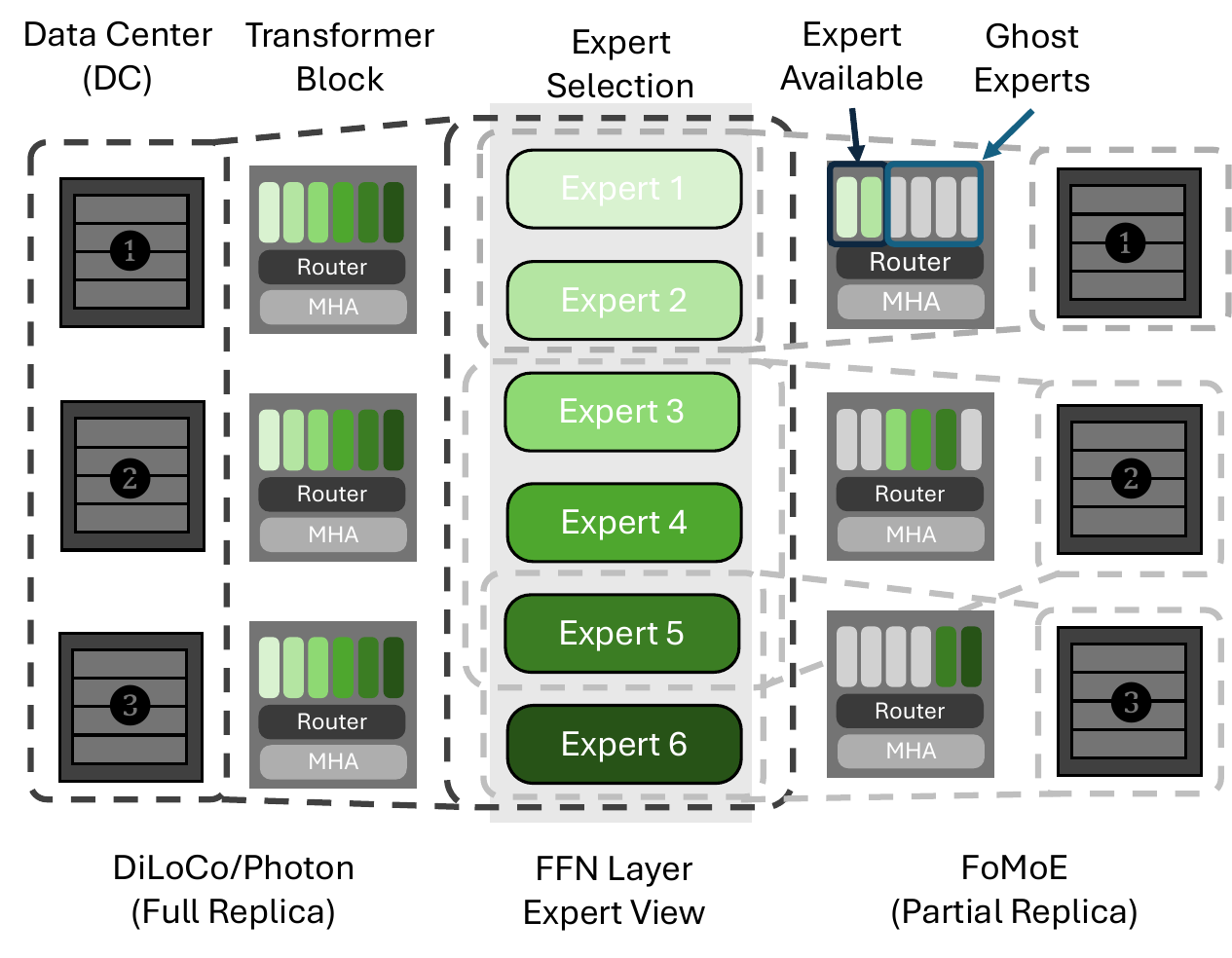}
    \caption{\method breaking the ``full-replica'' barrier for \wan-connected \moe training. Methods such as \diloco and \photon maintain a full \moe replica at each site and communicate this object at round boundaries. \method instead leverages \moe sparsity so each worker trains and communicates only the experts assigned to it. This design targets cross-site \wan settings where activation dispatch is too expensive.}
    \label{fig:teaser}
    \vspace{-0.3cm}
\end{figure}

State-of-the-art Large Language Models (\llms) continue to grow in scale, driven by the relationship between parameter count and model quality~\cite{kaplan2020scalinglaws,hoffmann2022chinchilla}. Training frontier \textit{dense} Transformer models has therefore become increasingly expensive, pushing compute, memory, and high-bandwidth interconnect limits in specialized training clusters~\cite{hoffmann2022chinchilla,rajbhandari2020zero,athlur2022varuna,palak2024geotraining}. To address this computational bottleneck, the field has increasingly adopted the \emph{Mixture-of-Experts} (\moes) architecture~\cite{shazeer2017moe, fedus2021switch}. By activating only a \textit{sparse} subset of parameters per token, models like \texttt{DeepSeek-V2/V3}~\cite{liu2024deepseekv2,deepseekv3} partially decouple computational cost from model capacity. This modularity reduces per-token compute, but it does not remove the need for fast links between hardware accelerators: conventional \moe training dispatches token activations to active experts during the forward and backward passes, so those experts are typically co-located in large data centers with high-bandwidth interconnects. As a result, \moe pre-training remains poorly matched to ordinary Internet links, volunteer or consumer-device pools~\cite{ryabinin2020towards,ryabinin2023swarm}, and other settings that could pool accelerators across sites rather than within a single tightly coupled cluster.

To aggregate more compute, frontier labs have shown interest in training across geographically distributed, weakly connected sites, effectively imitating a single supercomputer \cite{comanici2025gemini25pushingfrontier,Guthrie_2025,epoch2025coulddecentralizedtrainingsolveaispowerproblem}. For example, Google's Gemini $2.5$ was trained using connected data centers~\cite{comanici2025gemini25pushingfrontier}. However, the current understanding of this training paradigm remains in its infancy. In search of more efficiency in such cross-datacenter training, algorithms and systems like \diloco or \photon have emerged that reduce bandwidth overhead by limiting the synchronization \emph{frequency} during training~\cite{douillard2023diloco, sani2024photon}. This makes WAN-connected training less bandwidth-hungry, but it leaves a second bottleneck untouched: how much model state must move per round and how much model memory each site must hold.

However, a critical inefficiency remains: these methods typically treat the model as a \textbf{dense, monolithic replica} and do not reduce \textbf{payload size}. This approach forces every participating site to store the entire model state and exchange full parameter payloads, which often amount to hundreds of gigabytes, at every synchronization step. Consequently, cross-datacenter training faces two prohibitive barriers: (1) \textbf{Bandwidth}: full-payload transfers create massive latency spikes on \wan links, rendering training extremely challenging for truly large models, and (2) \textbf{Memory}: the maximum model size is capped by the storage capacity of the most memory-constrained site. This creates a mismatch: while modern \moe models possess \textbf{sparse data paths}, their distributed training infrastructure remains \textbf{communication-dense} and \textbf{memory-inefficient}. We argue that leveraging \textbf{communication sparsity} complements the computational efficiency of architectures like \texttt{DeepSeek}, enabling the pooling of \textbf{large-scale distributed resources} and further closing the gap between open and proprietary models.

In this work, we present \textbf{\method}, a cross-site training system that breaks the \textbf{``full-replica'' barrier} by co-designing the training strategy with the model's structural sparsity. Rather than maintaining identical replicas at every site, \method partitions each expert layer and assigns subsets of expert modules to different data centers. As a result, \method recovers the bandwidth-efficiency properties of previous methods \cite{douillard2023diloco, sani2024photon} by reducing synchronization frequency, but also reduces the \emph{payload size} sent during each synchronization. Since a worker does not fetch missing experts over the \wan during local training, \method also requires a path for tokens routed to non-resident experts; our skip-token mechanism bypasses those expert paths rather than dispatching activations across sites.

Relative to data-center approaches such as expert parallelism (EP), \method moves a different object across the network. EP shards experts within a high-bandwidth cluster and sends token activations to remote experts during every forward and backward pass~\cite{rajbhandari2022deepspeedmoe,hwang2022tutel,gale2022megablocks,liu2024deepseekv2,deepseekv3}. \method assumes workers never exchange token activations over the \wan; they train with local experts and synchronize selected model state only at round boundaries. EP can still be used within a site, while \method targets the cross-site payload and memory bottleneck left by full-replica low-communication training.

Our evaluation studies a controlled setting with homogeneous workers, i.i.d.\ data partitions, and controlled \wan bandwidth and latency. This isolates the effect of expert partitioning and payload reduction from orthogonal concerns such as asymmetric links, jitter, stragglers, and stale synchronization.

\method enables communication-efficient training of \moe models across multiple data centers without sacrificing machine learning performance. Our contributions are:

\begin{compactenum}
    \item \textbf{The \method System.} We introduce a framework for cross-site \moe training that supports partial expert replication. We define the design space of \emph{partitioning} (balancing memory vs.~bandwidth) and \emph{placement} (mapping experts to workers), exploring both fixed and random strategies to trade off expert specialization against generalization.
    \item \textbf{Partitioned FFN Optimization.} We demonstrate that partial expert replication linearly reduces communication volume, outperforming communication-efficient baselines by up to $1.42 \times$ and \texttt{DDP} by up to $45.44 \times$ in communication costs while maintaining perplexity in our controlled static-link evaluation. This shifts the empirical non-dominated trade-off frontier, enabling ML performance at lower system costs in the studied regimes.
    \item \textbf{``Skip-Token'' Throughput Gains.} We introduce the skip-token mechanism required by partial expert placement: when a token routes to a non-local expert (a ``ghost expert''), the worker bypasses that expert path instead of sending the token over the \wan. We show that this improves throughput linearly in theory and achieves empirical speedups up to \textbf{1.4$\times$}.
    \item \textbf{Scaling and Stability.} We separate the evidence by scale: trained proxy configurations provide empirical routing and convergence evidence, while the larger configurations in \cref{tab:model_configs} are analyzed through communication, memory, \flop, and wall-clock system models. Our results show that \method maintains high routing entropy and avoids expert collapse in the trained regimes, and that the same partial-replication mechanism projects favorable payload and memory scaling under the modeled large-scale configurations.
\end{compactenum}
\vspace{+0.1em}
\noindent \method bridges the gap between high-bandwidth, tightly-coupled \moe training and low-bandwidth, loosely-coupled distributed paradigms. By reducing the payload size per synchronization, it complements existing low-frequency methods~\cite{douillard2023diloco, sani2024photon}, paving the way for training next-generation \llms on globally sourced compute infrastructure.

\section{Background}\label{sec:background}

\noindent\textbf{Large-scale model training.}
Distributing computation is essential for large-scale model training. Standard data parallelism replicates the full model and synchronizes gradients per step \cite{li2020pytorch}, often saturating interconnects \cite{narayanan2019pipedream}. For memory constraints, model parallelism partitions the network: pipeline parallelism divides layers into sequential stages \cite{huang2019gpipe,narayanan2019pipedream}, while tensor parallelism splits layers across devices \cite{shoeybi2019megatron}. The \emph{Zero Redundancy Optimizer} (\texttt{ZeRO}), a form of data parallelism, eliminates memory redundancy by sharding parameters and optimizer states \cite{rajbhandari2020zero}. While these methods assume high-bandwidth fabrics, \emph{Varuna} adapts pipeline parallelism to commodity cloud \vms with limited \wan bandwidth \cite{athlur2022varuna}.

\vspace{+0.3em}
\noindent\textbf{\moe for scalable \llms.}
Mixture-of-Experts (\moe) architectures decouple parameter count from compute cost by activating sparse expert \texttt{MLPs} per token \cite{shazeer2017moe}. Implementations like \texttt{GShard} (Top-2) and \emph{Switch Transformers} (Top-1) achieve trillion-parameter capacities efficiently \cite{lepikhin2021gshard,fedus2021switch}. Recent advances reinforce this efficiency (\texttt{GLaM} \cite{du2022glam}) and optimize architectural design (\texttt{DeepSeek-V2/V3} \cite{liu2024deepseekv2,deepseekv3}). Theoretical works unify \moe perspectives via $\sigma$-\moe \cite{csordas2023sigma} and establish fine-grained scaling laws \cite{krajewski2024fgmoe}. Alternative routing strategies, including \texttt{BASE} layers \cite{lewis2021base}, \texttt{Expert-Choice} \cite{zhou2022expertchoice}, and \texttt{Hash Layers} \cite{roller2021hash}, address load balancing, while \texttt{V-MoE} addresses vision \cite{riquelme2021vmoe}. Expert pruning and skipping methods remove or bypass low-utility experts to improve deployment efficiency \cite{lu2024notall}; \method uses different skip-token semantics during training, in which missing remote experts are deliberately not executed across the \wan.

\vspace{+0.3em}
\noindent\textbf{Systems support for \moe and expert parallelism.}
Commodity \moe training leverages specialized runtimes (e.g., \texttt{DeepSpeed-MoE}, \texttt{Tutel}, \texttt{MegaBlocks}, \texttt{FasterMoE}, \texttt{SmartMoE}) to optimize All-to-All communication, memory management, and kernel execution \cite{rajbhandari2022deepspeedmoe, hwang2022tutel, gale2022megablocks, he2022fastermoe, zhai2023smartmoe}. These systems typically rely on \emph{expert parallelism} (EP): experts are sharded across devices, while token activations are dispatched to the owning experts over a fast intra-\dc fabric. EP can reduce per-device memory and enable large \moe within a cluster, but it does not remove the need for high-bandwidth activation exchange during each forward and backward pass. DeepSeek-style node-limited routing narrows this dispatch scope within a cluster~\cite{deepseekv3}; \method instead treats the \wan as a hard boundary, avoids cross-site activation dispatch entirely, and communicates only selected model state at synchronization boundaries. Thus, our design is orthogonal to intra-\dc EP and targets the payload and memory overheads that remain in full-replica cross-site training.

\vspace{+0.3em}
\noindent\textbf{Low-bandwidth distributed training.}
Training across weakly connected sites builds upon distributed optimization and local-update foundations \cite{mcmahan2017fedavg, reddi2020fedopt, li2020fedprox, stich2018localsgd}. Standard data parallelism synchronizes a full model every step; \diloco \cite{douillard2023diloco} and \photon \cite{sani2024photon} instead perform multiple local steps before each synchronization, reducing the frequency-dependent bandwidth cost by roughly the local-step interval. Theoretically, \cite{cheng2025localadam} proves convergence for adaptive optimizers in this regime, provided optimizer states are synchronized. Further communication reductions are achieved by decoupling model components (\texttt{DEPT} \cite{iacob2024dept}), desynchronizing optimizer states (\texttt{DES-LOC} \cite{iacob2025desloc}), co-designing modular paths (\texttt{DiPaCo} \cite{douillard2024dipaco}), or combining local updates with compression and sparse peer communication (\texttt{SparseLoCo}, \texttt{DiLoCoX}, \texttt{GASLoC}) \cite{sarfi2025sparseloco,qi2025dilocox,cagnasso2026gasloc}. Related systems work studies geo-distributed and weakly connected training through workload-aware WAN scheduling, pipeline parallelism, and volunteer-device \moe, including Atlas/BubbleTea, SWARM Parallelism, and Learning@home/DMoE \cite{palak2024geotraining,ryabinin2023swarm,ryabinin2020towards}. However, a direct mapping of these methods to \moe pre-training still materializes the full dense backbone and all experts at each site, or targets different axes such as pipeline scheduling, peer topology, or compression. \method addresses the complementary dimension of \emph{which} \moe experts must be present and synchronized at each site, thereby reducing per-round payload size and steady-state memory, rather than only the synchronization frequency.

\section{\method: The Federated MoE System}\label{sec:systemdesign}

Although current low-communication pre-training methods reduce the frequency of synchronization, and thus the cumulative communication cost, they still require transmitting full parameter payloads during updates~\cite{douillard2023diloco, sani2024photon}.
The straightforward adoption of \moe architectures in this setting achieves a \flop reduction for each worker by leveraging the structural \textbf{sparsity} of the model -- only $k \ll E$ of the total number of experts ($E$) are activated per token~\cite{fedus2021switch, liu2024deepseekv2, krajewski2024fgmoe, shazeer2017moe}. However, when applied directly to methods such as \photon or \diloco, the communication costs remain unchanged compared to dense models. By design, these methods require that the full parameter payload be communicated at every synchronization step, despite the inherent sparsity observed in these \moe architectures.

To alleviate this issue, we propose a federated \moe framework tailored to bandwidth-constrained wide-area network environments (\wan), leveraging the \textbf{sparsity} present in \moe architectures to reduce \textbf{both computational and communication costs} by \textbf{only training, and thus communicating, a subset of experts in each worker}.

We now present the \method system and its design space, covering: (a) \textbf{model partitioning}, which determines the optimal degree of expert replication; (b) \textbf{expert placement}, which orchestrates the distribution of experts across workers; and (c) \textbf{machine learning optimization}, which ensures stability and performance parity with high-communication baselines. \Cref{fig:teaser} shows a simplified system view of \method.

\subsection{Architecture and design}

Our system architecture is optimized for communication-constrained federated environments.
By breaking the ``full-replica'' barrier, we decouple the model size from synchronization costs. 
We detail the setting below and in \cref{sec:systemdesign:partition}.

\vspace{+0.4em}
\noindent{\textbf{Architecture and distributed setting.}}  
We target cross-datacenter training environments characterized by a distinct bandwidth hierarchy typical of real-world deployments: high intra-datacenter bandwidth $B_{\text{intra}}$ and severely constrained cross-datacenter bandwidth $B_{\text{cross}}$~\cite{palak2024geotraining}. We explicitly focus on the regime where $B_{\text{intra}} \gg B_{\text{cross}}$; the utility of local training strategies decreases when bandwidths are comparable ($B_{\text{intra}} \approx B_{\text{cross}}$). Formally, we consider a set $W$ of homogeneous data-parallel workers $M=|W|$, where each worker functions as a datacenter-class node with accelerators interconnected at $B_{\text{intra}}$ (e.g., $\approx$ \, \SI{200}{\giga\bit\per\second}). Each worker samples i.i.d.\ data partitions and connects to peers through \wan links with effective bandwidth $B_{\text{cross}}$ (e.g., $\approx$\,\SI{1}{\giga\bit\per\second}). In this setting, standard computation-communication overlap offers minimal benefit as transfer times dominate the latency. To mitigate this, we perform $K$ local optimization steps between fully synchronous rounds \cite{douillard2023diloco, sani2024photon}. We employ a central coordinator to manage initialization, broadcasting, and aggregation, restricting our study to synchronous training. We leave asynchronous extensions to future work due to the complexity of managing stale updates~\cite{douillard2025streaming} and note that these methods are fully \textbf{compatible} and \textbf{orthogonal} to our work.

\subsection{Model partitioning design}\label{sec:systemdesign:partition}

Although our system can flexibly accommodate arbitrary \moe architectures \cite{liu2024deepseekv2,kimiteam2025kimik2openagentic}, we fix the architecture design in this work. Specifically, our system trains decoder-only Transformer models \cite{vaswani2017attention} in which we replace the standard feed-forward network (\ffn) with $\sigma$-\moe-\ffn modules~\cite{csordas2023sigma,krajewski2024fgmoe} whose expert and router parameters are, \emph{potentially}, only \textbf{partially} replicated across \textbf{subsets} of workers (we provide full details of the architecture design in \cref{app:model_architecture}). Our architectural choice also supports capturing shared knowledge among experts~\cite{liu2024deepseekv2}.
\noindent Conceptually, \method categorizes the model parameters $\theta$ into the following.
\vspace{+0.3em}
\begin{compactenum}
    \item \textbf{Dense components} ($\theta_{\text{dense}}$): \textit{Fully replicated} parameters including attention weights, layer norms, shared experts, and input/output embeddings (tied by default).
    \item \textbf{Sparse components} ($\theta_{\text{sparse}}$): \textit{Potentially partitioned} parameters including expert weights and routing networks. We detail the worker partitioning and assignment of sparse components in \cref{subsec:experts_partitioning,sec:systemdesign:placement}.
\end{compactenum}
\vspace{+0.3em}
To illustrate this breakdown, we consider a typical Transformer configuration. The token embedding matrix comprises $d_{\text{model}} \times |\mathcal{V}|$ parameters where $\mathcal{V}$ is the vocabulary. Each attention layer contains four weight matrices totaling $4 d_{\text{model}}^2$ parameters (assuming standard head dimensions). In contrast, an \moe layer adds a routing network with $d_{\text{model}} \times N_e$ parameters (where $N_e$ is the total expert count) and the experts’ \ffns. For a dense expansion ratio $e$, we allocate a total \ffn hidden dimension of $e \cdot d_{\text{model}}$ per layer, divided among $N_e$ experts (each with hidden size $d_{\text{ff,expert}} = \frac{e\,d_{\text{model}}}{N_e}$). Consequently, the combined expert parameter count is $e \times d_{\text{model}}^2$, comparable to a dense \ffn.

\begin{figure}[htb]
    \centering
    \includegraphics[width=\linewidth]{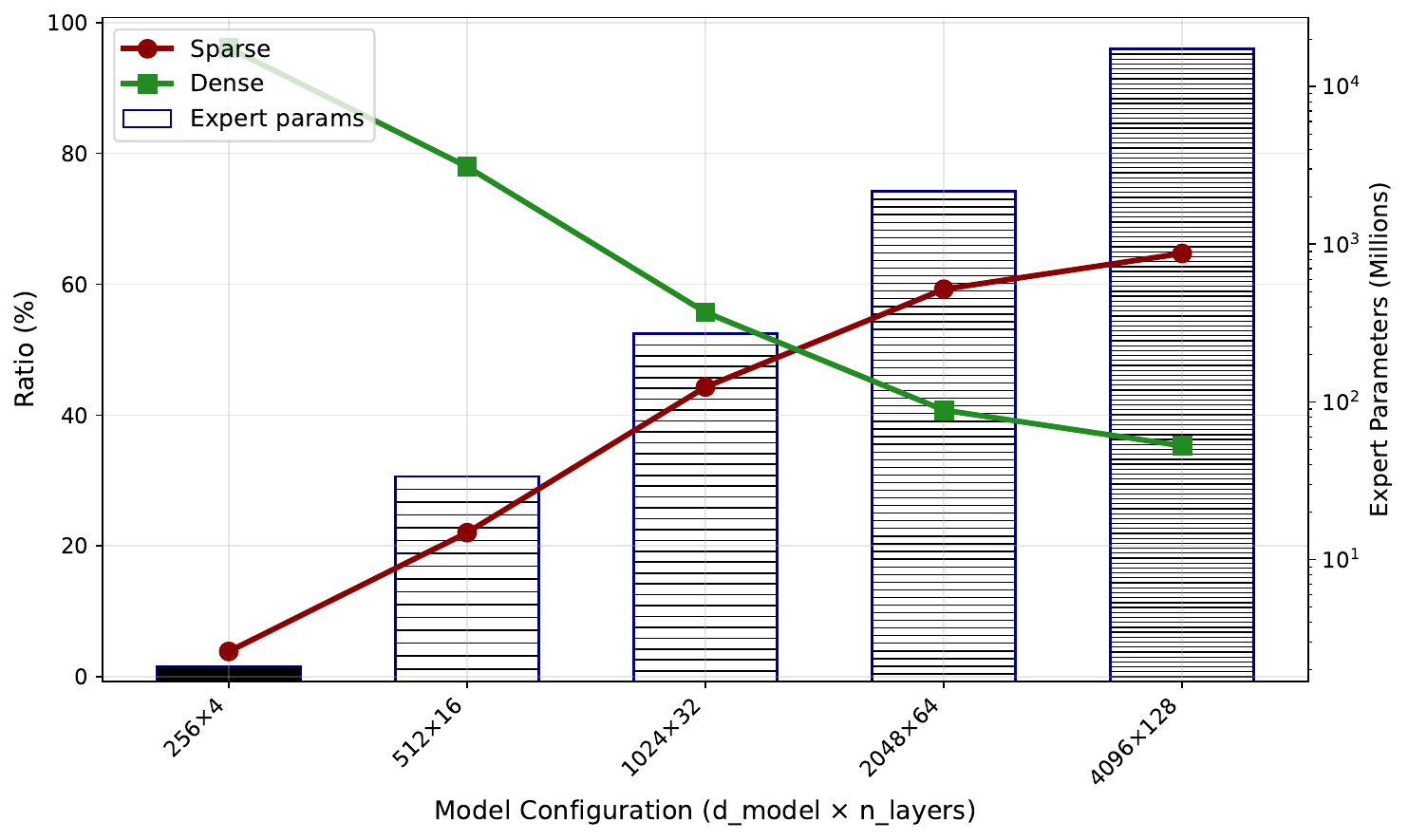}
    \caption{Parameter distribution trends. Scaling width ($d_{\text{model}}$) dramatically increases the proportion of sparse parameters ($\theta_{\text{sparse}}$), motivating the focus of our strategy.}
    \label{fig:parameter_distribution_scaling}
\end{figure}

As model scale increases, the ratio $\theta_{\text{sparse}}/\theta$ shifts significantly. While scaling depth primarily adds dense parameters, scaling width (i.e., increasing $d_{\text{model}}$) primarily adds experts. Adopting a standard width-to-depth ratio of $d_{\text{model}} : L \approx 32$~\cite{dey2025completep}, $\theta_{\text{sparse}}$ grows from $\approx 20\%$ of total parameters at the $150$M scale ($d_{\text{model}}=512$) to $>60\%$ at the $26$B scale ($d_{\text{model}}=4096$), as shown in \cref{fig:parameter_distribution_scaling}. This relationship shows that the efficient synchronization of $\theta_{\text{sparse}}$ is the primary lever for enhancing communication efficiency as model size increases.
This scaling design is well established in the literature \cite{csordas2023sigma,malasnicki2025muparam,krajewski2024fgmoe} and facilitates the study of the broader design space of expert placement options as the number of expert modules increases.

\begin{widefigure}[ht!]
    \centering
    \includegraphics[width=\linewidth]{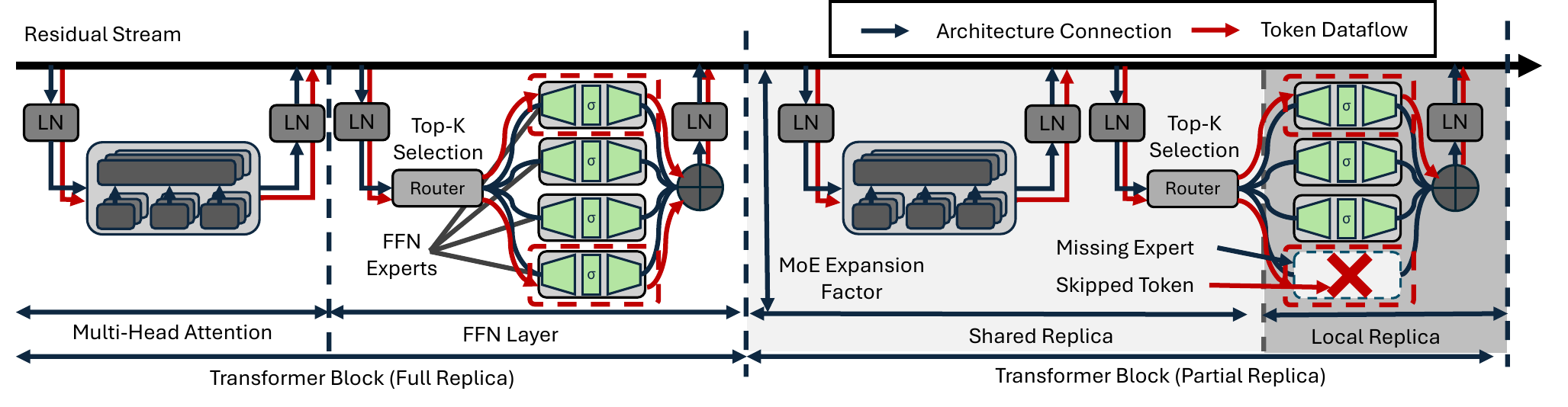}
    \caption{Comparison between a transformer \moe block as a full versus partial replica. In the partial-replica form, the same experts can be selected per token as in the full-replica case, but missing experts are ignored by the local computational graph, yielding savings in \flop, communication, and memory.}
    \label{fig:skipped-token}
\end{widefigure}

\subsection{Partitioning and placement strategies}

\method enables the distribution of sparse parameters controlled via two coupled design decisions: \textbf{partitioning}, which determines expert replication and balances memory and bandwidth; and \textbf{placement}, which maps experts to workers. These mechanisms jointly control communication costs and effective model capacity. We define participation in \cref{subsec:experts_partitioning} and the placement algorithms in \cref{sec:systemdesign:placement}.

\subsubsection{Experts partitioning and scalability}\label{subsec:experts_partitioning}

To analyze the impact of partitioning at scale, consider a layer $\ell$ with $N_{e,\ell}$ experts distributed across $M$ workers. We introduce the \emph{overlapping factor} $\mathcal{O}_e \in \{1,\ldots,M\}$, defined as the number of distinct workers hosting any specific expert. This parameter controls the design spectrum from completely disjoint partitions ($\mathcal{O}_e=1$) to full replication ($\mathcal{O}_e=M$). Consequently, each worker hosts a reduced local subset of $N_{\text{le},\ell} = \frac{\mathcal{O}_e}{M} \, N_{e,\ell}$ experts. During training, local routing is bounded by $\min(k,\,N_{\text{le},\ell})$ experts (where $k=2$ by default). Partitioning ($\mathcal{O}_e < M$) yields threefold advantages over full-replicas methods:
\vspace{+0.3em}
\begin{compactenum}
    \item \textbf{Communication:} the payload scales with the fraction of globally shared parameters $\mathcal{O}_e/M$, saving the remaining $1 - \frac{\mathcal{O}_e}{M}$ from cross-site transfer.
    \item \textbf{Compute:} computation can be reduced in two ways. First, the router scores only $N_{\text{le},\ell}$ experts, reducing router \flops by $\approx \mathcal{O}_e/M$. Second, if $N_{\text{le},\ell} < k$, the system activates fewer experts per token, thus reducing \ffn \flops. Otherwise, the per-token compute matches standard \moe ($N_{\text{le},\ell} \ge k$).
    \item \textbf{Memory:} the per-worker memory footprint for expert weights decreases by a factor of $M/\mathcal{O}_e$, enabling training where na\"ive full replication would be too expensive, due to bandwidth and memory constraints.
\end{compactenum}

\subsubsection{Expert placement modes}\label{sec:systemdesign:placement}

\begin{algorithm}[t]
\caption{Expert Placement \& Migration Strategy}
\label{alg:expert_placement}
\footnotesize \begin{algorithmic}[1]
 \Require $M$ (workers), $N_e$ (total experts), $O_e$ (overlap)
 \Require $T$ (shuffle period), $t$ (round), $\mathcal{S}^{(t-1)}_m$ (prev. set)
 \Require \texttt{Mode} (Fixed or Random)

 \State  \State $\mathcal{S}^{(t)}_m \gets \mathcal{S}^{(t-1)}_m$ \hfill \textcolor{gray}{\scriptsize // Default: retain set}

 \State \textbf{\textit{Phase 1: Placement Logic}}
 \If{$t = 0$ \textbf{or} (\texttt{Mode} = Rand \textbf{and} $t \pmod T = 0$)}
    \State $\mathcal{P} \gets \texttt{GenPlacement}(N_e, O_e, M)$
    \State $\mathcal{S}^{(t)}_m \gets \texttt{GetLocalSubset}(\mathcal{P}, m)$
    \State \textcolor{gray}{\scriptsize $\triangleright$ Assigns $N_{le} = \frac{O_e}{M}N_e$ experts to worker $m$}
 \EndIf

 \State \textbf{\textit{Phase 2: Migration \& State Mgmt.}}
 \State $\Delta_{\text{in}} \gets \mathcal{S}^{(t)}_m \setminus \mathcal{S}^{(t-1)}_m$ \hfill \textcolor{gray}{\scriptsize // Identify new experts}
 
 \If{$\Delta_{\text{in}} \neq \emptyset$}
    \State $\texttt{Broadcast}(\Delta_{\text{in}})$ 
    \State \textcolor{gray}{\scriptsize $\triangleright$ Fetch missing weights (prob. $p_{miss} \approx 1 - \frac{O_e}{M}$)}
    
                 \Else
    \State \textcolor{gray}{\scriptsize $\triangleright$ Maintain local momentum (no migration)}
 \EndIf
 
 \State \textbf{return} $\mathcal{S}^{(t)}_m$
\end{algorithmic}
\end{algorithm}

Once the expert partitioning configuration (defined by $\mathcal{O}_e$ and $N_{e,\ell}$) is set, \method enables flexible expert-to-worker assignment. We investigate two modes, detailed in \cref{alg:expert_placement}:

\vspace{+0.3em}
\noindent\textbf{Fixed placement.} Upon initialization of the $M$ workers, the central server deterministically maps each expert in layer $\ell$ to $\mathcal{O}_e$ workers. Consequently, each worker is assigned a static set of $N_{\text{le},\ell}$ experts. These assignments remain fixed throughout training, with each worker responsible solely for updating its local experts. In the limiting case where $\mathcal{O}_e=1$, this scheme reduces to disjoint, fully local experts; workers collaborate exclusively on the dense parameters $\theta_{\text{dense}}$. This effectively specializes each site to a specific subset of expert modules, analogous to decoupled training methods such as \dept~\cite{iacob2024dept}.
While DeepSeek's Expert Parallelism (EP) treats the cluster as a unified memory pool by dynamically sharding experts to maximize capacity~\cite{liu2024deepseekv2}, \method's fixed placement treats the \wan as a hard constraint, strictly pinning experts to specific workers to enforce data locality and eliminate weight synchronization during local steps.

\vspace{+0.3em}
\noindent\textbf{Random placement.} To improve expert generalization, this mode reassigns experts to workers at a cadence of $T$ synchronization rounds. The server randomly redistributes experts subject to the constraint that every expert must reside on exactly $\mathcal{O}_e$ workers, and every worker must host $N_{\text{le},\ell}$ \textbf{distinct} experts. Under this scheme, the specific expert set resident on a worker evolves over time. A small reshuffling period ensures experts are trained across worker data partitions, potentially enhancing generalization. However, this introduces trade-offs: it incurs network overhead for migrating model states and may affect local optimizer momentum (which we discuss in \cref{sec:systemdesign:comm}). In practice, the reshuffling period is tuned to balance data mixing against stability and costs. \Cref{eq:sparse_bytes_cost} demonstrates that while reshuffling can, in the worst case, necessitate migrating every expert, this overhead decreases linearly as $\mathcal{O}_e \to M$ and as the reshuffling interval increases relative to the communication period.

\noindent\textbf{System context and routing logic.}
We emphasize that, in our federated setting, token routing is \textbf{strictly local}: tokens processed at a site are dispatched exclusively to experts hosted at that site, with no inter-site exchange of activations. This design maximizes locality, avoiding high-latency \wan communication during the forward and backward passes. This distinguishes \method from systems like \texttt{DeepSpeed-MoE}, \texttt{Tutel}, \texttt{MegaBlocks}, \texttt{FasterMoE}, \texttt{SmartMoE}, which optimize execution \emph{within} high-bandwidth clusters via all-to-all communication~\cite{rajbhandari2022deepspeedmoe,hwang2022tutel,he2022fastermoe,zhai2023smartmoe}.
Our placement strategies thus operate at a distinct granularity from typical intra-layer load balancing. While we assume standard mechanisms (e.g., \texttt{BASE} layers~\cite{lewis2021base}, \texttt{Expert-Choice}~\cite{zhou2022expertchoice}, or capacity scaling~\cite{fedus2021switch}) are employed to balance loads among \emph{available} local experts, our work addresses the \wan-constrained challenge of determining \emph{which} experts are available at each site.

\noindent \textbf{Placement trade-off.} Random expert placement may outperform fixed placement when data or hardware heterogeneity makes additional expert mixing valuable. First, while a fixed topology restricts a worker's router to a static support set of $N_{\text{le}}$ experts, limiting the routing combinations to $\binom{N_{\text{le}}}{k}$, random placement reshuffles assignments every synchronization interval. Over $R$ rounds, a worker explores up to $\min ( R, \binom{N_e}{N_{\text{le}}} )$ distinct support sets. Second, randomization can act as a regularizer when the i.i.d.\ assumption fails, since it decouples experts from specific data shards and exposes them to a broader range of data distributions. In homogeneous i.i.d.\ settings evaluated in \cref{sec:eval:placement}, however, the migration and optimizer-state costs dominate this hypothesized mixing benefit.

\subsubsection{``Ghost Experts'' and ``Skip-Token''}

For most of our work, we partition each router, along with the experts, so that individual tokens are not routed to experts that are unavailable to the local worker.
However, we investigate the opportunity to go beyond such biased routing and to further leverage the sparsity and partitioning of experts, reducing computation.
We propose an alternative partitioning, as seen in \cref{fig:skipped-token}, that does not partition the routing matrices, allowing each router to assign individual tokens to experts who are not locally available to the workers.
The cost of not partitioning the routing matrices and communicating them in full, as if they are part of the dense components $\theta_{\text{dense}}$, is negligible compared to that of the expert modules.
We refer to those experts that can receive token assignments but are not available as ``ghost experts''.
The token assignments to the ``ghost experts'' provide an opportunity to further reduce \flops, since we cannot follow the computational path of these tokens due to missing expert weights.
When the router's assignments are perfectly balanced, the executed expert \flops scale with the ratio of local experts to total experts, so the skipped computation grows with the complementary ghost-assignment fraction.
This should be interpreted as an approximation whose quality impact must be measured, not as an assumption that the router has a large top-$1$/top-$2$ score margin or that experts are non-redundant.
In the present experiments, we do not use router-score margins to justify skipping; instead, we evaluate the downstream effects directly via validation perplexity and routing stability in \cref{fig:exp1_skip_tokens}.
In contrast to DeepSeek's dispatch mechanism, which relies on high-bandwidth interconnects to route tokens to remote experts for full computation~\cite{liu2024deepseekv2}, \method uses its ``ghost expert'' strategy to skip computation for remote experts, trading a potential reduction in active expert computation for the complete elimination of cross-node activation transfer.

\begin{figure}[htb]
    \centering
    \includegraphics[width=\linewidth]{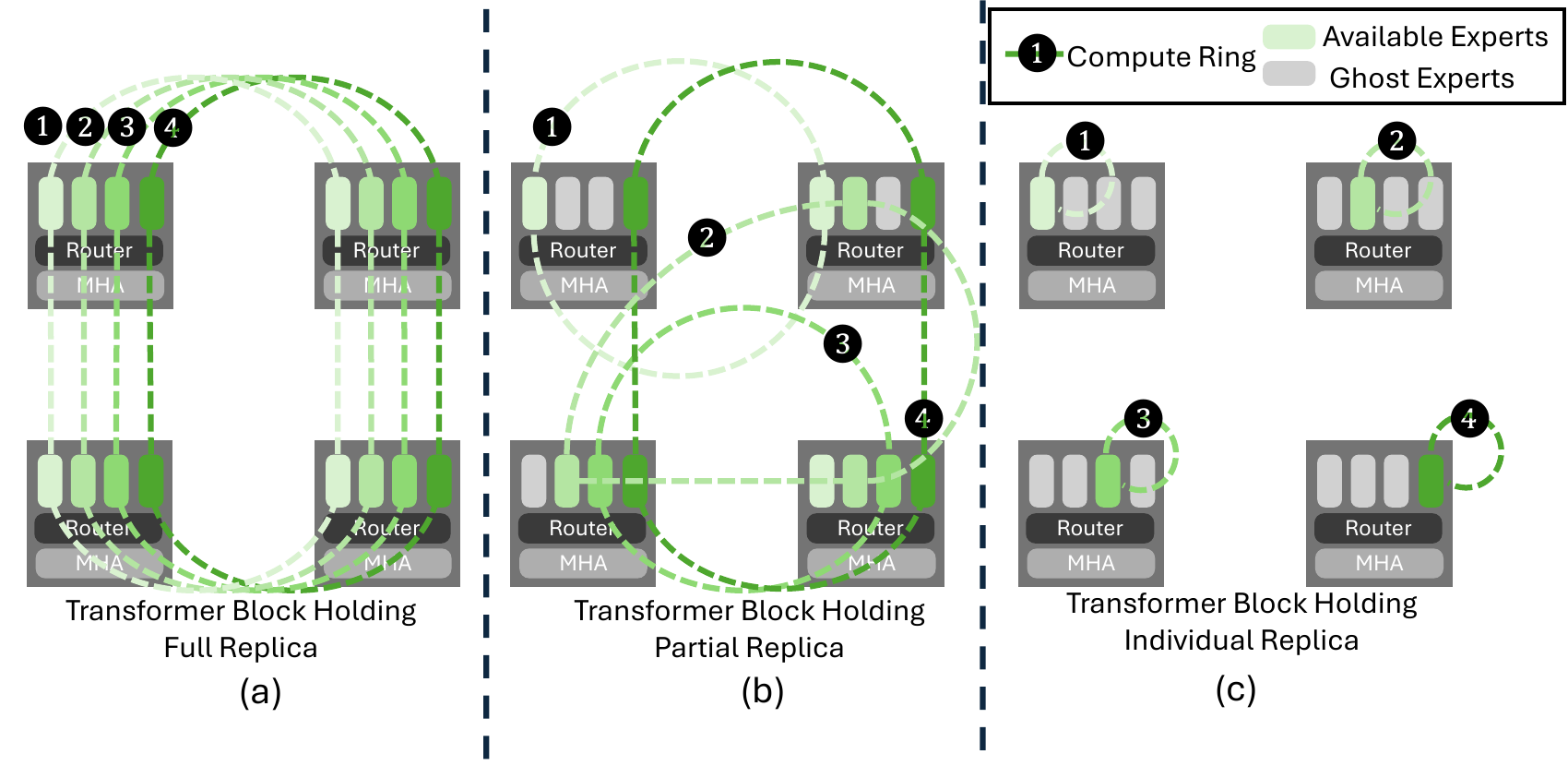}
    \caption{Expert placement and ring synchronization. In a full-replica regime, each available expert incurs its own full ring across workers. As $O_e \to 1$, the communication cost $R(O_e)$ decreases because each expert ring spans fewer workers; at $O_e=1$, each ring is localized to the worker that owns the expert.}
    \label{fig:expert-placement}
\end{figure}

\subsection{Synchronization and model optimization}\label{sec:systemdesign:comm}

\method training (as we show in \cref{alg:simplified_fomoe}) proceeds in rounds~\cite{mcmahan2017fedavg}: the server broadcasts parameters, workers perform $K$ local steps (this implies a synchronization frequency reduction of $1/K$), and updates are aggregated. We make the following choices regarding the training process.

\vspace{+0.3em}
\noindent\textbf{Round structure and collectives.}
We utilize bandwidth-optimal \texttt{Ring-AllReduce} for aggregation. Averaging a tensor of size $S$ across $M$ peers entails transmitting $\frac{2(M-1)}{M}S$ bytes per peer~\cite{rabenseifner2004reduce,patarasuk2009allreduce}. While we employ a flat reduction, the framework supports hierarchical strategies for heterogeneous networks~\cite{kairouz2021advances}. Furthermore, asynchronous techniques like \texttt{Streaming} \diloco~\cite{douillard2025streaming} can overlap communication with computation; while we have not implemented such overlapping in \method, it can be readily incorporated into our performance modeling (e.g., as a compression factor).

\vspace{+0.3em}
\noindent\textbf{Outer optimization.}
We combine infrequent synchronization ($K \gg 1$) with adaptive optimization. Workers execute \adamw~\cite{loshchilov2019adamw} locally on their parameter subsets, while the server applies momentum (\fedopt) to the aggregated model to maintain convergence~\cite{reddi2020fedopt,khaled2025outer}. Crucially, the local optimizer states are preserved rather than reset. For the global $\theta_{\text{dense}}$, states are tracked by all workers; for $\theta_{\text{sparse}}$, workers maintain history only for their own experts.

A challenge arises in the \emph{random placement} regime, where expert migration separates weights from their optimizer history. To avoid the overhead of transferring states when experts are reassigned, we reinitialize the optimizers for reassigned experts with a learning-rate warmup applied only to the reassigned weights. In contrast, \emph{fixed placement} continuously retains the state. Empirically, we observe that preserving local momentum prevents the divergence observed in stateless approaches. To ensure scalability without retuning, we transfer the following hyperparameters: global learning rate schedules (leveraging \mup/\completep~\cite{yang2021tensorprogram,dey2025completep,malasnicki2025muparam} transfer), \adamw settings ($\beta_1, \beta_2, \epsilon$), warmup duration, and router temperature. See \cref{sec:systemdesign:transfer} for more details.

\noindent\textbf{Bandwidth regimes and trends.}
Unlike high-speed intra-datacenter settings (\qtyrange[range-phrase=~--~]{100}{400}{\giga\bit\per\second}) where communication overlaps computation~\cite{rajbhandari2022deepspeedmoe}, \wan constraints (\qtyrange[range-phrase=~--~]{1}{20}{\giga\bit\per\second}) necessitate minimizing volume ($\mathcal{O}_e \ll M$) and frequency ($K \gg 1$). Although increasing $M$ or $K$ exacerbates model drift, the server momentum effectively stabilizes training~\cite{stich2018localsgd}. Recent works confirm that with tuned hyperparameters, infrequent synchronization maintains perplexity parity~\cite{douillard2023diloco,sani2024photon} while reducing communication overhead~\cite{iacob2025desloc}. Gradient compression~\cite{alistarh2017qsgd} remains orthogonal.
\section{\method Scalability \& Resource Consumption}\label{sec:scalability-costs}

Building on the system design in \cref{sec:systemdesign}, we define a quantitative cost model to optimize \method configurations and hyperparameters under resource constraints.
Given $M$ weakly connected sites with effective \wan bandwidth $B_{\mathrm{eff}}$ and fixed hardware, we determine the expert overlap $\mathcal{O}_e$, synchronization interval $K$, and expert count $N_{e}$ that maximize performance per wall-clock time. These models help determine how to set up \method for the specific requirements of the underlying training infrastructure.

The model comprises three components: (1) computational cost based on \flops (\cref{sec:compute-memory}); (2) communication cost based on data volume (\cref{subsec:comm-scaling}); and (3) a wall-clock time model synthesizing these factors within the distributed environment (\cref{sec:walltime}).
This framework identifies the partitioning methodology of \cref{sec:systemdesign}, synchronization interval $K$, expert overlap factor $\mathcal{O}_e$, and worker count $M$ as the primary levers.
While prior work \cite{douillard2023diloco,sani2024photon} examined general communication-computation trade-offs, they neglected \moe-specific partitioning.
Our model explicitly captures the trade-offs of partial expert replication in bandwidth-constrained multi-datacenter settings, demonstrating how tuning $\mathcal{O}_e$ and $K$ overcomes the ``full-replica'' barrier to efficient distributed training.

\subsection{Compute and Memory Scaling}\label{sec:compute-memory}

Training a dense Transformer with $N$ parameters on $D$ tokens requires approximately $6ND$ \flops \cite{kaplan2020scalinglaws}, obeying power-law scaling between compute and model quality \cite{hoffmann2022chinchilla}. Sparse models \moe partially decouple the parameter count from the compute cost by activating only $k \ll N_e$ experts per layer. The effective parameter count for compute becomes $N_{\text{dense}} + \phi N_{\text{exp}}$; $\phi = k/N_e$ is the active fraction \cite{du2022glam} of the total expert parameters. Since $\phi \ll 1$, \moes achieve a trillion-parameter scale with per-token cost of significantly smaller dense models \cite{riquelme2021vmoe,shazeer2017moe,lepikhin2021gshard,fedus2021switch}.

In our federated topology, local compute costs depend on the available experts. If a worker holds $N_{\text{le},\ell} \ge k$ experts, the costs match that of full-replica training. However, if $N_{\text{le},\ell} < k$ (e.g., each worker has a single expert when $k=2$), the worker activates only $\min(k, N_{\text{le},\ell})$ experts, reducing \ffn \flops \cite{fedus2021switch}. Notably, the gating network projects only to local experts, scaling router operations by $N_{\text{le},\ell}/N_{e,\ell} = \mathcal{O}_e/M$. Reducing expert overlap $\mathcal{O}_e$ lowers per-token compute, especially when forcing sparse activation (e.g., top-1) \cite{gale2022megablocks}.

We estimate the per-token \flops $F_{\text{token}}$ for a worker with overlap $\mathcal{O}_e$ and local expert count $N_{\text{le},\ell} = \frac{\mathcal{O}_e}{M}N_{e,\ell}$ as:
\begin{equation}\label{eq:flops_per_token} \resizebox{0.90\hsize}{!}{$
F_{\text{token}} \;\approx\; F_{\text{dense}} \;+\; \sum_{\ell=1}^L \Bigg(\frac{\min(k,\,N_{\text{le},\ell})}{k}\,F_{\text{FFN},\ell} \;+\; \frac{N_{\text{le},\ell}}{N_{e,\ell}}\,F_{\text{gate},\ell}\Bigg)\,$
}
\end{equation}
where $F_{\text{dense}}$, $F_{\text{FFN},\ell}$, and $F_{\text{gate},\ell}$ denote the per-token \flops of the dense components, unpartitioned expert \ffns, and full gating networks, respectively.

In the presence of ``ghost experts'' adopting the ``skip-token'' procedure, the formula requires applying a scaling factor that accounts for the possibility of skipping the \ffn computation for a token.
With a perfectly balanced routing network, such a scaling factor can be written as:
\begin{equation}\label{eq:skip_token_scaling}
\sigma_{\text{st}} = \frac{N_{\text{le},\ell}}{N_{e,\ell}} \le 1
\end{equation}
and, consequently, the total flops per token can be expressed, more generally, as:
\begin{equation}\label{eq:flops_per_token_st}
\resizebox{0.90\hsize}{!}{$F_{\text{token}} \;\approx\; F_{\text{dense}} \;+\; \sum_{\ell=1}^L \Bigg(\sigma_{\text{st}}\frac{\min(k,\,N_{\text{le},\ell})}{k}\,F_{\text{FFN},\ell} \;+\; \frac{N_{\text{le},\ell}}{N_{e,\ell}}\,F_{\text{gate},\ell}\Bigg)\,$}
\end{equation}
We note that routing decisions are not always perfectly balanced; introducing a load-balancing loss ensures they converge to a balanced state.
In the presence of a severe imbalance in the assignment, the scaling factor above should be replaced with the actual frequency of the token assignments to ``ghost experts'' against the total number of assignments to provide more precise cost estimation.
In our experiments, we observe an overall balance, with the empirical scaling factor always being very close to its ideal counterpart.
We recommend running a few warm-up iterations in full-replica mode to detect excessive imbalance before it becomes too problematic to maintain machine learning performance.

\vspace{+0.3em}
\noindent\textbf{Memory Footprint.}
Large-scale training memory comprises \emph{steady-state} allocations for parameters and optimizer states, and \emph{dynamic} allocations for activations and temporary buffers \cite{rajbhandari2020zero,micikevicius2018mixed}. A standard mixed-precision \adamw requires $\approx$ \SI{18}{\byte} per parameter ($2$ for \texttt{FP16} weights, $4$ for \texttt{FP32} master weights, $8$ for moments, and $4$ for gradients). Activation memory scales with batch size and sequence length, and activation checkpointing can trade additional compute for lower activation storage \cite{chen2016checkpointing}.

Our \method architecture reduces both memory components. Partitioning experts across $M$ sites with overlap $\mathcal{O}_e$ reduces the stored expert parameters per site to a fraction $\frac{\mathcal{O}_e}{M}$. The expected per-site steady-state memory is approximately $\frac{\mathcal{O}_e}{M}\sum_{\ell,e} s_{\ell,e} + S_{\text{dense}}$, where $s_{\ell,e}$ and $S_{\text{dense}}$ denote the size of expert and dense parameters, respectively \cite{rajbhandari2022deepspeedmoe}. Reducing $\mathcal{O}_e$ toward $1$ yields nearly linearly proportional memory savings, particularly for models where sparse experts dominate the parameter count (e.g., $>60\%$ in \texttt{DeepSeek-V2} \cite{liu2024deepseekv2}). Furthermore, fewer local experts reduce the dimensionality of the gating network, providing additional steady-state savings.

Dynamic memory also decreases with lower $\mathcal{O}_e$: when a worker holds fewer experts than the global top-$k$ (i.e., $N_{\text{le},\ell} < k$), it computes activations for only $\min(k, N_{\text{le},\ell})$ experts per token. A local top-$1$ execution in a global top-$2$ regime halves the stored \ffn activation tensors: fewer local experts reduce activation memory and gradient buffers. Such savings are additive to standard memory-optimization techniques \cite{fedus2021switch,rajbhandari2020zero}.

\subsection{Communication Scaling}\label{subsec:comm-scaling}

We quantify the communication overhead by deriving the \emph{bytes-per-token} (\texttt{BPT}) exchanged between sites. This metric directly links the synchronization interval $K$ and the expert overlap $\mathcal{O}_e$ to \wan bandwidth requirements.

\vspace{+0.3em}
\noindent\textbf{Setting and Aggregation.}
As detailed in \cref{alg:simplified_fomoe}, training proceeds in synchronous rounds where $M$ sites perform $K$ local steps before aggregating parameters \cite{mcmahan2017fedavg,reddi2020fedopt,li2020fedprox}. We partition the model into fully replicated dense parameters $\theta_{\text{dense}}$ and sparse expert weights $\{\theta_{\text{sparse}}\}$ distributed according to the overlap factor $\mathcal{O}_e$ (where $\mathcal{O}_e$ denotes the number of workers hosting a specific expert).
Various strategies modify the synchronization payload and frequency: \localsgd \cite{stich2018localsgd} and \diloco \cite{douillard2023diloco} sync only model parameters, keeping optimizer states local; \localadamw \cite{cheng2025localadam} syncs both, roughly tripling traffic; and \desloc \cite{iacob2025desloc} decouples parameter and state synchronization schedules, enabling flexible trade-offs in parameter vs optimizer state synchronization. Outer-optimizer analyses further explain the roles of learning rates, momentum, and acceleration in local-update regimes \cite{khaled2025outer}. We consider a subset of these strategies for our parameter-partitioned \moe regime.

\vspace{+0.3em}
\noindent\textbf{Modeling Communication Costs.}
Synchronizing fully replicated components ($\theta_{\text{dense}}$ of size $S_{\text{dense}}$ in bytes) via \texttt{Ring-AllReduce }\cite{rabenseifner2004reduce,patarasuk2009allreduce} incurs a per-site cost of:
\begin{equation}\label{eq:dense_bytes_cost}
B^{\text{per-site}}_{\text{dense}}(M) = \frac{2(M-1)}{M}\,S_{\text{dense}}\,
\end{equation}
For sparse components, communication depends on the overlap $\mathcal{O}_e$. Assuming a uniform assignment where each worker holds $N_{\text{le},\ell} = \frac{\mathcal{O}_e}{M}N_{e,\ell}$ experts, the per-site communication cost depend on two steps: (1) synchronization and aggregation of experts' weights which can be performed with \texttt{Ring-AllReduce}; and (2) the broadcast due to potential re-shuffling of the experts across workers (only occurs for the random placement strategy).
During step (1), a worker will need to exchange data for each local expert with a cost weighted by a ring factor:
\begin{equation}\label{eq:experts_ring_factor}
R(\mathcal{O}_e) = \frac{2(\mathcal{O}_e - 1)}{\mathcal{O}_e}
\end{equation}
Notably, when $\mathcal{O}_e = 1$, this cost disappears since no synchronization and aggregation occur.
During step (2), the communication cost is determined by the set difference between the experts assigned at round $t+1$ and those already present at round $t$. We model this as a probabilistic process.

Assuming a uniform random assignment, the probability that a specific expert $e$ assigned to a worker in the new placement was already present in its local partition $S_t$ is equal to the fraction of global experts held locally:
\begin{equation}    
p_{\text{hit}} = \frac{N_{\text{le},\ell}}{N_{e,\ell}} = \frac{(\frac{\mathcal{O}_e}{M} N_{e,\ell})}{N_{e,\ell}} = \frac{\mathcal{O}_e}{M}
\end{equation}

The probability that an assigned expert is missing and requires migration (a ``broadcast'' event) is $p_{\text{miss}} = 1 - \frac{\mathcal{O}_e}{M}$.

We define the weighting factor $P(\mathcal{O}_e, M)$ as this miss probability, scaled by the reshuffling frequency $\gamma$ (where $\gamma = 1$ implies reshuffling every synchronization round, and $\gamma < 1$ implies amortized periodic reshuffling). 
As discussed in \cref{sec:systemdesign:placement}, we investigate two placement modes, fixed and random, which correspond to the setting $\gamma=0$ and $\gamma=1$, respectively.
Substituting into \cref{eq:experts_broadcast}:

\begin{equation}\label{eq:experts_broadcast}
P(\mathcal{O}_e, M, \gamma) = \gamma \left( 1 - \frac{\mathcal{O}_e}{M} \right)
\end{equation}

This derivation aligns with the boundary conditions discussed in \cref{subsec:comm-scaling}:
\begin{compactitem}
    \item \textbf{Full Replica ($\mathcal{O}_e = M$):} The term becomes $1 - 1 = 0$. Since workers hold all the experts, no migration is ever required, and the synchronization term also accounts for the cost of weight updates.
    \item \textbf{Disjoint Partition ($\mathcal{O}_e = 1$):} The term becomes $1 - \frac{1}{M} \approx 1$. A worker holds a negligible fraction of the global experts, so nearly every reshuffling event requires replacing the local expert set entirely.
\end{compactitem}

Adding these components together, the communication cost due to sparse components of the model is:
\begin{equation}\label{eq:sparse_bytes_cost}
\resizebox{0.9\hsize}{!}{$
B^{\text{per-site}}_{\text{sparse}} (\mathcal{O}_e, M, \gamma) = (R(\mathcal{O}_e) + P(\mathcal{O}_e, M, \gamma))\;\sum_{\ell}\Big(N_{le,\ell}\,s_{\text{expert},\ell}\Big)\,$}
\end{equation}
where $N_{le,\ell}$ is the number of local experts, $s_{\text{expert},\ell}$ is the single expert size in bytes, which includes up and down projection, and its router projector if not using the ``skip-token'' procedure.
This formulation recovers the full-replica cost when $\mathcal{O}_e=M$ (matching \cref{eq:dense_bytes_cost}).
Notably, with $\mathcal{O}_e=1$, the system requires one worker to act as an orchestrator to handle the reassignment of distinct sparse components that are not shared among workers.
In such cases, the ring factor $\frac{2(\mathcal{O}_e -1)}{\mathcal{O}_e}$ disappears, and the cost uniquely depends on the summation. Strategies synchronizing optimizer states (e.g., \localadamw) multiply $B_{\mathrm{AR}}^{\text{per-site}}$ by a constant factor (typically $3\times$) \cite{cheng2025localadam}.

\vspace{+0.3em}
\noindent\textbf{Bytes-Per-Token (\texttt{BPT}).}
For a site processing $T_{\text{local}} = K B_l$ tokens per round, where $K$ is the number of local optimization steps and $B_l=\frac{B_g}{M}$ is the local batch size (global batch size $B_g$ split across the $M$ parallel workers), the total \texttt{BPT} is derived from the sum of dense and sparse costs ($B^{\text{per-site}} = B^{\text{per-site}}_{\text{dense}} + B^{\text{per-site}}_{\text{sparse}}$):
\begin{equation}
\mathrm{BPT} \;=\; \frac{B^{\text{per-site}}}{K B_l}
\;=\; \frac{M B^{\text{per-site}}}{K B_g}
\end{equation}
This relation highlights three key levers: increasing $K$ or batch size $B$ amortizes communication; minimizing $\mathcal{O}_e$ caps the sparse term.

\subsection{Wall-Clock Time Model}\label{sec:walltime}

We synthesize the computational (\cref{sec:compute-memory}) and communication (\cref{subsec:comm-scaling}) models to estimate the total training time $T_{\mathrm{tot}}$. This unified framework serves as a ``Roofline'' model for federated \moe training, identifying bottlenecks to guide the selection of the synchronization interval $K$, the expert overlap $\mathcal{O}_e$, and the reshuffling frequency $\gamma$.

\vspace{+0.3em}
\noindent\textbf{Computational Time ($t_{\mathrm{comp}}$).}
The local optimization step time, $t_{\mathrm{comp}}$, depends on the dynamic \flop count and hardware efficiency. Unlike dense baselines, $t_{\mathrm{comp}}$ scales with the partitioning strategy. Let $\Phi_{\mathrm{peak}}$ be the theoretical peak \flops and $\mathrm{MFU}(\cdot)$ be the Model \flops Utilization, which is sensitive to local batch size $B_l = \frac{B_g}{M}$ and arithmetic intensity of the reduced expert kernels. Using $F_{\text{token}}(\mathcal{O}_e)$ from \cref{eq:flops_per_token_st}:
\begin{equation}
    t_{\mathrm{comp}}(\mathcal{O}_e, M) = \frac{F_{\text{token}}(\mathcal{O}_e) \cdot \frac{B_g}{M}}{\mathrm{MFU}(\frac{B_g}{M}, N_{\text{le},\ell}) \cdot \Phi_{\mathrm{peak}}}\,
\end{equation}
Reducing $\mathcal{O}_e$ lowers $F_{\text{token}}$ due to skipping tokens, reducing $t_{\mathrm{comp}}$. However, this is bounded by the fragmentation limit: if the local expert count $N_{\text{le},\ell}$ becomes too small relative to $B_l$, $\mathrm{MFU}$ degrades due to insufficient tensor core saturation.

\vspace{+0.3em}
\noindent\textbf{Communication Time ($t_{\mathrm{comm}}$).}
Communication time is governed by the effective \wan bandwidth $B_{\mathrm{eff}}$. While intra-datacenter interconnects (e.g., NVLink) offer 100--400 Gbps \cite{rajbhandari2022deepspeedmoe}, cross-site \wan links typically operate at 1--20 Gbps \cite{jaghouar2024opendiloco}, making transfer time bandwidth-dominated. We extend the payload formulation to account for gradient compression (e.g., QSGD \cite{alistarh2017qsgd}), which is an orthogonal addition to our work, via a factor $\rho \in (0, 1]$, which is $\rho=1.0$ by default:
\begin{equation}
\resizebox{0.85\hsize}{!}{
    $t_{\mathrm{comm}}(M, \mathcal{O}_e, \gamma) \approx \frac{\rho \cdot \left[ B^{\text{per-site}}_{\text{dense}}(M) + B^{\text{per-site}}_{\text{sparse}}(\mathcal{O}_e, M, \gamma) \right]}{B_{\mathrm{eff}}}\,$
}
\end{equation}
This term captures the trade-off of our strategy: lowering $\mathcal{O}_e$ minimizes gradient synchronization (the aggregation cost), but utilizing random placement ($\gamma > 0$) introduces a state migration penalty (the broadcast factor $P \to \gamma$).

\vspace{+0.3em}
\noindent\textbf{Total Training Time.}
The total time $T_{\mathrm{tot}}$ for a token budget of $N_{\mathrm{tok}}$ sums the costs of $K$ compute steps and one communication step per round, subject to an overlap factor $\alpha$. With global batch size $B_g = M \cdot B_l$ and total rounds $\frac{N_{\mathrm{tok}}}{K \cdot B_g}$:
\begin{equation}\label{eq:master_equation}
\resizebox{0.85\hsize}{!}{
$ T_{\mathrm{tot}} \approx \frac{N_{\mathrm{tok}}}{M \cdot B_l \cdot K} \left[ K \cdot t_{\mathrm{comp}}(M, \mathcal{O}_e) + (1-\alpha) t_{\mathrm{comm}}(M, \mathcal{O}_e, \gamma) \right]\,$}
\end{equation}

\noindent\textbf{Model Dynamics.}
This framework highlights three critical dynamics captured by the model:

\begin{compactenum}
    \item \textbf{The Federated Compute-Bound Regime:} Ideally, training satisfies $K \cdot t_{\mathrm{comp}} \gg (1-\alpha) t_{\mathrm{comm}}$. If inter-site bandwidth is comparable to intra-site, partitioning offers minimal benefit. However, on commodity \wan, full-replica approaches ($\mathcal{O}_e=M$) may require excessively large $K$ to satisfy this inequality. \method achieves this regime with moderate $K$ by aggressively reducing payload size via partial replication ($\mathcal{O}_e \to 1$). This improvement is multiplicative with other optimizations; for example, maximal overlap ($\alpha \to 1$) and compression ($\rho \ll 1$) can make low-bandwidth training more attainable even for $100$B+ modeled configurations on \SI{1}{\giga\bit\per\second} links~\cite{qi2025dilocox}.

    \item \textbf{The Memory-Time Coupling:} \cref{eq:master_equation} reveals a quadratic efficiency gain from reducing $\mathcal{O}_e$. Lowering overlap reduces memory pressure (\cref{sec:compute-memory}), enabling a larger local batch size $B_l$. Since the communication overhead ratio is proportional to $\frac{1}{K \cdot B_l}$, increasing $B_l$ dilutes the fixed communication cost over more tokens. Thus, reducing $\mathcal{O}_e$ linearly reduces the payload ($B^{\text{per-site}} \downarrow$) increases throughput per sync ($B_l \uparrow$).

    \item \textbf{The Placement Convexity:} Under random placement, the interaction between aggregation and migration costs dictates an optimal overlap $\mathcal{O}_e^*$. If reshuffling is frequent ($\gamma \approx 1$), the migration penalty makes disjoint partitioning ($\mathcal{O}_e=1$) suboptimal. \method targets a ``Semi-Static'' regime ($\gamma \ll 1$) to amortize migration costs, allowing the system to reduce bandwidth via disjoint experts.
\end{compactenum}

Solving \cref{eq:master_equation} for a specific hardware profile ($B_{\mathrm{eff}}, \Phi_{\mathrm{peak}}$) yields the theoretical optimal configuration $(\mathcal{O}_e^*, K^*)$ that minimizes wall-clock time. In practice, we generate a set of hardware-efficient configurations ranked by increasing training time and empirically evaluate their task performance (e.g., language modeling loss). This sweep optimizes the trade-off between system efficiency and task performance, aiming to maximize performance per time unit.

\section{Experimental Setting}\label{sec:experimental-setting}

We evaluate \method across a range of model scales to quantify trade-offs among communication efficiency, computational throughput, and machine learning quality. The experiments focus on the modeling and cost effects of partial expert replication: convergence, routing stability, communication volume, and throughput. They deliberately avoid characterizing every production-system concern, such as time-varying jitter, asymmetric links, stragglers, or heterogeneous accelerators.

\vspace{+0.3em}
\noindent\textbf{Model Configurations.} We define five model configurations: \textbf{Small Proxy (54M)}, \textbf{Medium (150M)}, \textbf{Large (2B)}, \textbf{XL (13B)}, and \textbf{XXL (100B)}. These span five operational regimes: rapid prototyping, entry-level modeling, compact open-weight scaling, standard open-weight deployment, and memory-bound scaling. All models utilize a decoder-only Transformer backbone as described in \cref{app:model_architecture}. All comparisons are made on an iso-parameter basis: each sparse \moe configuration matches the parameter count of its dense counterpart while reducing inference \flops by a factor of $0.25$. To configure the larger model sizes, internal components are scaled from the \textbf{Small Proxy (54M)} model, as in \cite{dey2025completep,malasnicki2025muparam,yang2021tensorprogram}. The specifics of the architectures are detailed in Table~\ref{tab:model_configs}.

\begin{table}[htbp]
    \centering
    \caption{Overview of the model configurations used in the ``ghost experts'' studies. Configurations range from a small proxy model to a massive $100$B parameter model, maintaining the same head dimension ($d_{head}=64$) and expansion factor ($4$).}
    \label{tab:model_configs}
    \resizebox{\columnwidth}{!}{    \begin{tabular}{lcccccccc}
        \toprule
        & \multicolumn{3}{c}{\textbf{Model Architecture}} & \multicolumn{3}{c}{\textbf{Model Configuration}} & \textbf{Scale} \\
        \cmidrule(lr){2-4} \cmidrule(lr){5-7} \cmidrule(lr){8-8}
        \textbf{Configuration} & $d_{model}$ & $n_{layers}$ & $n_{heads}$ & $N_{experts}$ & $d_{expert}$ & $N_{active}$ & \textbf{Total Params} \\
        \midrule
        Small Proxy & 256 & 4 & 4 & 8 & 128 & 2 & $\sim$54M \\
        Medium (150M) & 512 & 16 & 8 & 16 & 128 & 4 & $\sim$153M \\
        Large (2B) & 1024 & 128 & 16 & 32 & 128 & 8 & $\sim$1.8B \\
        XL (13B) & 2048 & 256 & 32 & 64 & 128 & 16 & $\sim$13.3B \\
        XXL (100B) & 4096 & 512 & 64 & 128 & 128 & 32 & $\sim$104B \\
        \bottomrule
    \end{tabular}    }
\end{table}

\noindent\textbf{Machine Learning Performance.}
We evaluate performance stability and convergence speed. To circumvent the prohibitive cost of hyperparameter tuning at the $100$B scale, we adopt the \textbf{\completep} parameterization framework~\cite{dey2025completep,malasnicki2025muparam,yang2021tensorprogram}. We first validate the optimal learning rates across cross-site configurations derived at the \textbf{Small Proxy (54M)} scale, and then transfer them to large-model settings. All models are trained on a large pre-training corpus comprising a mixture of general web-scraped text and code, inspired by the SmolLM Corpus~\cite{benallal2024smollmcorpus}. We adopted the GPTOss tokenizer for pre-processing the raw text~\cite{gptoss}. The training duration has been chosen to reach at least the Chinchilla compute optimality, i.e., $20$ tokens per parameter, and slightly adjusted to allow cross-site experiments to terminate at a round boundary~\cite{hoffmann2022chinchilla}. Subsequently, we conduct ablation studies on the overlapping factor $\mathcal{O}_e$ to quantify the resulting perplexity degradation due to reduced expert visibility. Finally, we analyze the routing dynamics of the ``skip-token'' mechanism to ensure that dropping tokens for ``ghost experts'' does not destabilize convergence relative to forced local routing. Each trained model was evaluated on the entire validation set of the \texttt{en} C4 corpus \cite{raffel2023exploringlimitstransferlearning}, and we report the mean and standard deviation of evaluation perplexity across the dataset.

\vspace{+0.3em}
\noindent\textbf{Communication Efficiency.}
We quantify network overhead using \textbf{Bytes Per Token (\texttt{BPT})} and total synchronization volume per round. We compare \method against full-replica baselines (e.g., \diloco/\photon), modeling the communication cost as a function of expert overlap. Theoretically, setting $\mathcal{O}_e=1$ (disjoint experts) restricts synchronization to the dense backbone ($\theta_{dense}$), so the payload reduction scales with the share of parameters in sparse experts. In the XL and XXL configurations of \cref{tab:model_configs}, where sparse experts dominate the model state, this removes most expert payload from each cross-site round.

\vspace{+0.3em}
\noindent\textbf{Computational Efficiency.}
Efficiency is measured via \textbf{Model \flops Utilization (\texttt{MFU})} and wall-clock throughput (tokens/sec). While standard partial replication maintains an iso-\flop count per token, the ``skip-token'' mechanism dynamically reduces the effective computational graph. We model the theoretical throughput of this mechanism and compare it with empirical results, identifying overheads due to router divergence and memory fragmentation.

\vspace{+0.3em}
\noindent\textbf{Model Size Scaling and Placement.}
We explore the high-dimensional design space defined by the interactions among model size ($N$), expert count ($N_e$), and cluster size ($M$). By varying the number of experts and workers, we analyze the fragmentation limit: the threshold where the reduction in local batch size per expert (due to high $N_e/M$ ratios) can degrade \texttt{GPU} tensor core saturation.

\vspace{+0.3em}
\noindent\textbf{Training Environment.}
Experiments are executed on GPU cluster nodes with NVIDIA accelerators. Unless a result is explicitly identified as an empirical throughput measurement, \wan bandwidth and latency values are treated as controlled cost-model parameters rather than a claim of packet-level network emulation. In particular, our modeled cross-site setting uses \SI{1}{\giga\bit\per\second} and \SI{10}{\giga\bit\per\second} links with \SI{50}{\milli\second} latency, without time-varying jitter, asymmetric bandwidth, or persistent stragglers. Since the present implementation is synchronous, a slow site would delay the corresponding synchronization round. Stale-synchronous or asynchronous variants, such as Streaming \diloco, are compatible with the cost model but left to future work.

\vspace{+0.3em}
\noindent\textbf{Implementation.}
We implement \method using PyTorch collective communication primitives and a lightweight coordination layer. The coordinator initializes dense parameters globally, assigns sparse expert subsets according to the placement policy, triggers local training for $K$ steps, and aggregates only the parameters owned by each expert replica group. Standard \texttt{Distributed Data Parallel} and \texttt{FSDP} assume static full-model replication or high-bandwidth intra-cluster sharding, so they cannot directly express the variable expert subset synchronization used by \method without an additional placement and aggregation layer.

\vspace{+0.3em}
\noindent\textbf{Reproducibility.}
\Cref{app:reproducibility} reports the full reproducibility summary for the experiments. The appendix records the training settings, sweep grids, dataset mixture, evaluation protocol, artifact status, and known gaps.

\section{Evaluation and Results}

We systematically evaluate \method to quantify the trade-offs between communication, throughput, and quality. The evidence combines three dimensions: trained small-proxy quality runs for convergence and routing stability, measured skip-token throughput on accelerator hardware, and analytical/modeled communication, \flop, memory, and wall-clock scaling for larger configurations.

\Cref{sec:exp0} benchmarks full-replica baselines and identifies which outer optimizer yields stronger communication-quality trade-offs than methods such as \diloco. \Cref{sec:eval:placement} and \cref{sec:eval:skip_tokens} evaluate expert overlap, placement, and token skipping. The main takeaway is that \method shifts the empirical Pareto frontier relative to \texttt{DDP} and full-replica, low-communication baselines, thereby providing improved performance at a given communication or compute cost in the regimes studied.
\Cref{sec:eval:scalability} evaluates modeled scalability, demonstrating that reducing expert overlap yields linear reductions in communication and wall-clock components under our cost model, while ``skip-token'' improves measured throughput without specialized kernels. To ensure that our baselines are well-tuned across all trained model scales, we adopt the \completep transfer framework \cite{dey2025completep,yang2021tensorprogram,malasnicki2025muparam}. We provide further context and details of the sweeps in \cref{sec:systemdesign:transfer}.

\subsection{Baseline Evaluation for Full Replica}
\label{sec:exp0}

\begin{figure}[ht]
    \centering
    \includegraphics[scale=0.4]{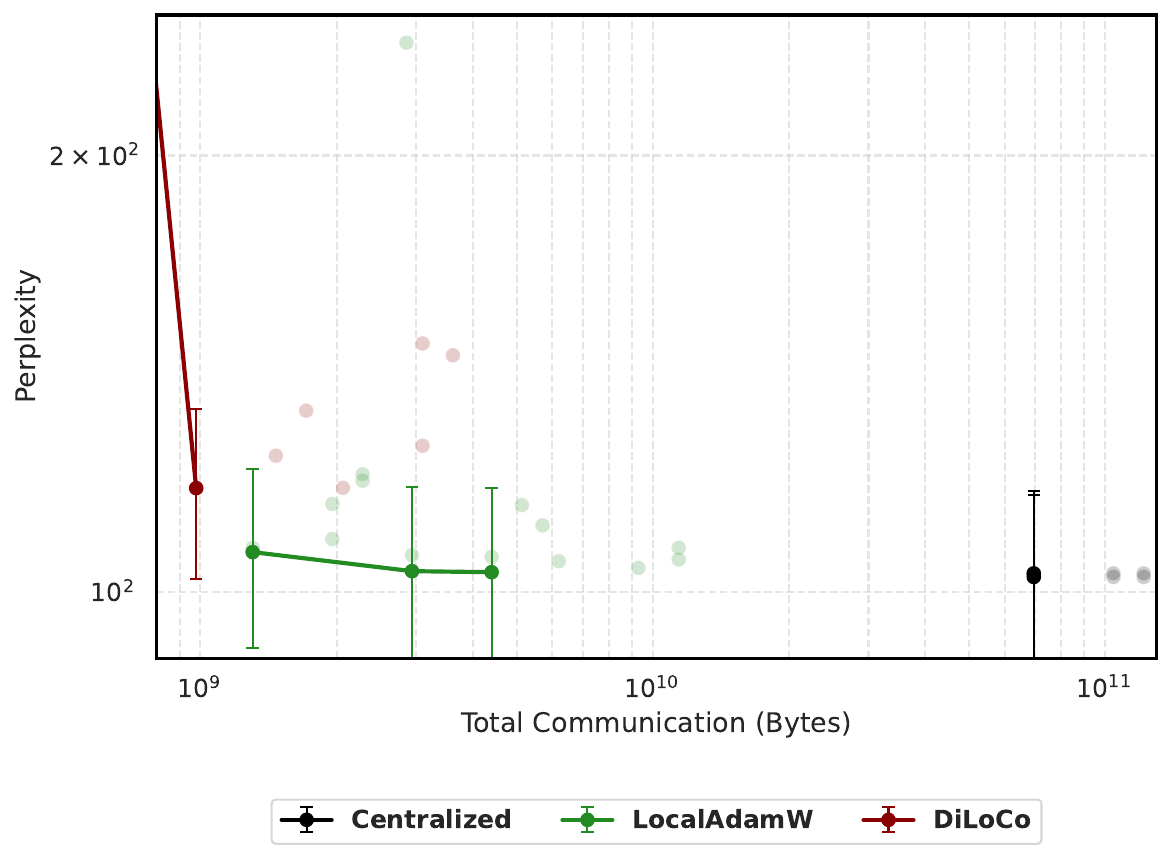}
    \caption{Pareto frontier of communication costs versus perplexity for full-replica baselines across outer optimizers. ``Centralized'' is the high-bandwidth single-site quality reference. \localadamw offers a better trade-off than \diloco within a given communication budget, motivating the use of the stateful optimizer in subsequent \method configurations.}
    \label{fig:exp0}
    \vspace{-0.1cm}
\end{figure}

We first calibrate two ``full replica'' baselines -- dense and \moe implementations -- on the perplexity-versus-communication Pareto front. Specifically, we evaluate full-replica runs across varying synchronization intervals $K \in \{32, 64, 128\}$ and worker counts $M \in \{2, 4, 8\}$, utilizing both \diloco and \localadamw outer optimizers. Experiments employ our \emph{small proxy} configuration (see \cref{tab:model_configs}) and an \emph{iso-token} budget, training all models to compute-optimal parameter counts. In these plots, \emph{centralized} denotes the single-site high-bandwidth reference training run with no WAN throttling or low-frequency cross-site synchronization; it is a quality reference and not a practical baseline. \emph{Full replica} denotes a cross-site configuration that keeps all dense and sparse parameters at every site and reduces only synchronization frequency. \cref{fig:exp0} therefore identifies the stronger full-replica outer-optimizer baseline used for the subsequent \method comparisons.

Taken only as baseline calibration, \cref{fig:exp0} identifies \localadamw as the stronger full-replica optimizer: despite the extra optimizer-state traffic, it tracks the centralized quality reference more closely than \diloco at comparable communication budgets. The appendix table in \cref{tab:exp0_extended_results} reports the full $K$ and $M$ grid and, consistent with prior work~\cite{sani2024photon,douillard2023diloco}, shows that increasing either value degrades perplexity through local drift and gradient noise. We therefore carry \localadamw forward as the main optimizer baseline for the \method trade-off comparison in \cref{fig:exp1}.

\subsection{\texorpdfstring{$\mathcal{O}_e$}{Oe} and Expert Placement Strategies}\label{sec:eval:placement}

\begin{figure}[ht]
\centering
        \includegraphics[width=0.68\linewidth]{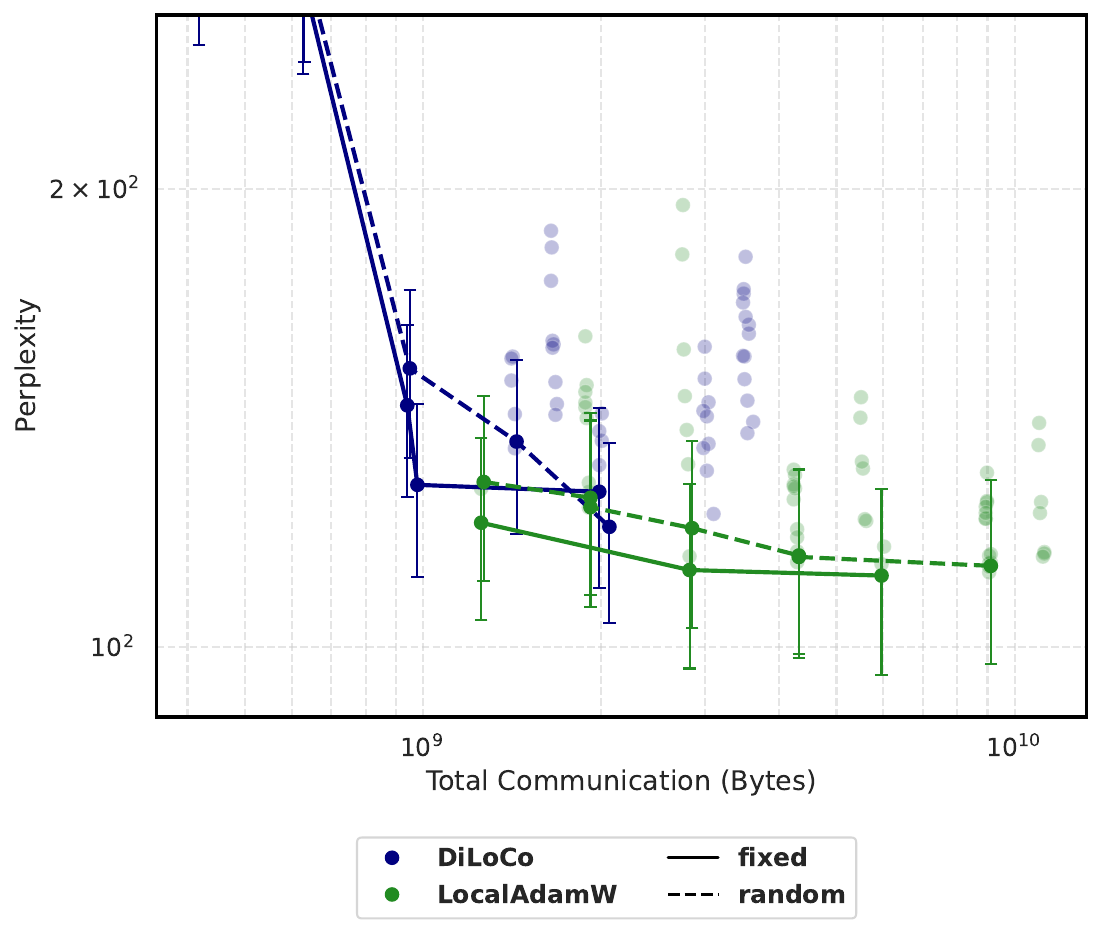}
        \vfill
        \includegraphics[width=0.68\linewidth]{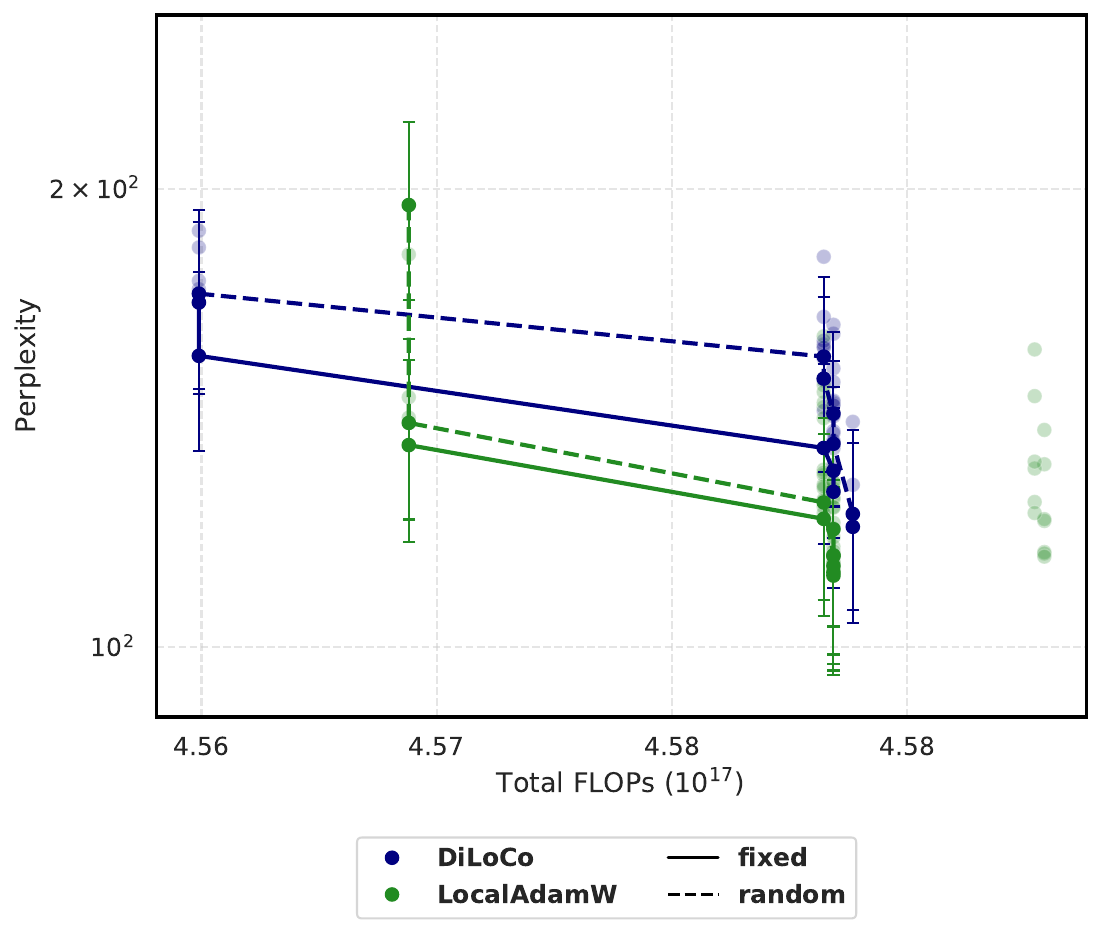}
    
    \caption{Empirical Pareto frontiers comparing communication cost (top) and computation cost (bottom) against perplexity for \method across different outer optimizers and placement strategies. In this sweep, the strongest observed non-dominated configurations occur with \localadamw and fixed expert placement. Momentum-vector averaging and expert reassignment affect the trade-off between regularization, data mixing, and additional migration overhead.}
    \label{fig:exp1}
    \vspace{-0.2cm}
\end{figure}

\cref{fig:exp1} gives the corresponding \method trade-off comparison, ablating the impact of $O_e$ and the placement strategy. Here, \localadamw yields lower-perplexity configurations with comparable communication budgets because synchronizing optimizer state improves stability in low-frequency synchronization regimes~\cite{cheng2025localadam,iacob2025desloc}. Across both \diloco and \localadamw, fixed placement yields the best configurations for both communication-quality and compute-quality trade-offs. Random placement does not shift this frontier in the homogeneous, i.i.d.\ sweep: additional mixing doesn't bring sufficient benefits considering the migration costs; under heterogeneous or non-i.i.d.\ data, where mixing may improve quality, this conclusion may change.

\noindent
Addressing the impact of $O_e$: across all combinations of synchronization intervals $K$, placement strategies, and workers $M$, we find that as $O_e \rightarrow M$, performance converges towards full-replica performance as seen in \cref{tab:exp1_results}. Smaller values of $O_e$ achieve higher perplexity under an iso-token training budget, as the models have less active capacity per site to capture the underlying data distribution. However, larger $O_e$ values incur higher communication and computation costs; $O_e \approx 4$ yields the configurations that best balance quality against these costs.

\subsection{``Skip-Token'' and ``Ghost Experts''}\label{sec:eval:skip_tokens}

\begin{figure}[htb]
    \centering
    \includegraphics[width=\linewidth]{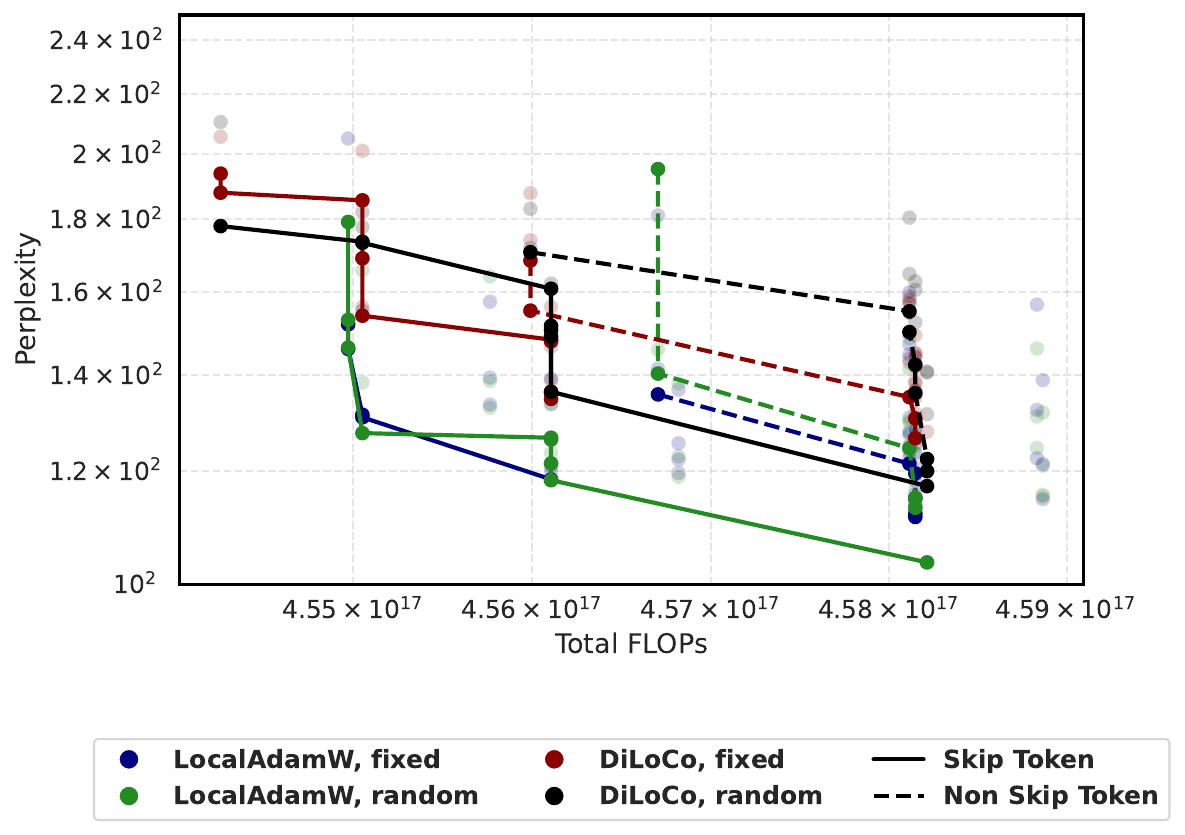}
    \caption{Impact of ``skip-token'' and ``ghost experts'' on the perplexity versus \flops Pareto frontier. Across outer optimizer configurations, these mechanisms shift the empirical compute-quality frontier toward lower \flop cost at comparable perplexity, and lower perplexity at comparable \flop cost, relative to full-replica baselines.}
    \label{fig:exp1_skip_tokens}
\end{figure}

We now determine the influence of the ``skip-token'' and ``ghost expert'' mechanisms on the empirical compute-quality frontier, as seen in \cref{fig:exp1_skip_tokens}.
Across both outer optimizers, introducing these mechanisms shifts the frontier by adding configurations with lower perplexity at comparable or lower compute budgets. This shift arises because each worker skips tokens routed to missing local experts rather than spending compute on unavailable expert paths. Additionally, these results show that adding the ``skip-token'' mechanism does not significantly impact training convergence.

\noindent\textbf{Quality interpretation.}
It is crucial to note that ``skip-token'' is not equivalent to full-replica training or to EP with remote activation dispatch, because it removes expert paths that are unavailable at the local site for a subset of the tokens in the batch. In this paper, we test its machine-learning effectiveness and stability through validation perplexity; here, routing stability should be interpreted as a supporting diagnostic. Future work must test whether, for bigger and more capable models, downstream task performance remains unchanged, especially for reasoning-heavy tasks where the router's ability to choose experts is critical, and assess longer training trajectories.

\subsection{\method at Scale}\label{sec:eval:scalability}

We assess \method's scalability using the theoretical framework from \cref{sec:scalability-costs}. We analyze \flops, communication costs, and wall-clock time as analytical/modeled quantities, not as full large-scale training runs. Results show that minimizing expert overlap ($O_e$) significantly reduces compute, communication, and time in the modeled configurations, enabling \method to outperform \texttt{DDP} and full-replica baselines under those assumptions.

\begin{figure}[htb]
    \centering
    {\includegraphics[width=0.49\linewidth]{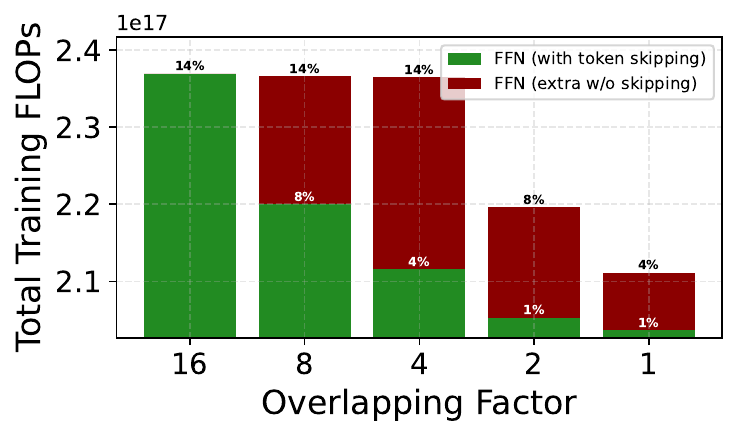}} \hfill
    {\includegraphics[width=0.49\linewidth]{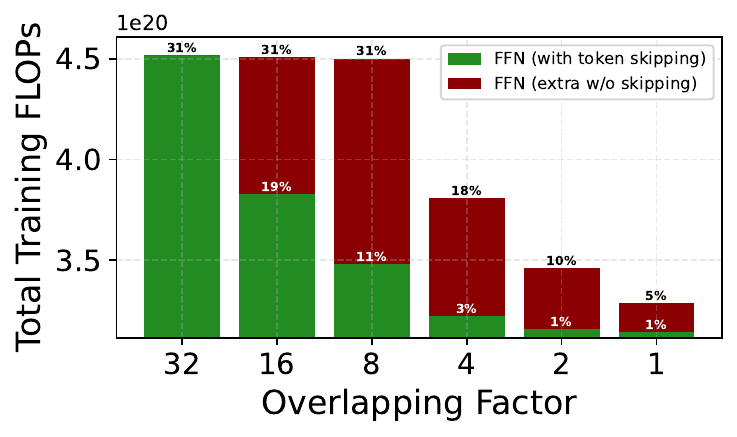}}
    \caption{\textbf{Analytical/modeled total training FLOPs versus expert overlap ($O_e$).}
    \textbf{Left:} Small model ($d_{model}=512$, 16 layers, 16 workers).
    \textbf{Right:} Large model ($d_{model}=8192$, 512 layers, 32 workers).
    Green bars show modeled FLOPs with token skipping; red bars indicate avoided computation. Lowering overlap reduces compute, with greater relative savings for larger models.}
    \label{fig:flops_scaling}
\end{figure}

\vspace{0.3em}
\noindent\textbf{\flops Scalability.}\label{eval:theory:flops}
\Cref{fig:flops_scaling} instantiates predictions from \cref{sec:scalability-costs}: (a) partitioning reduces router \flops, with linear \texttt{FFN} savings when $O_e < k$; (b) the ``skip-token'' mechanism further linearly reduces \flops by bypassing non-local experts; and (c) these savings grow with model scale as sparse parameters increasingly dominate.

\begin{figure}[htb]
    \centering
        {\includegraphics[width=.8\columnwidth]{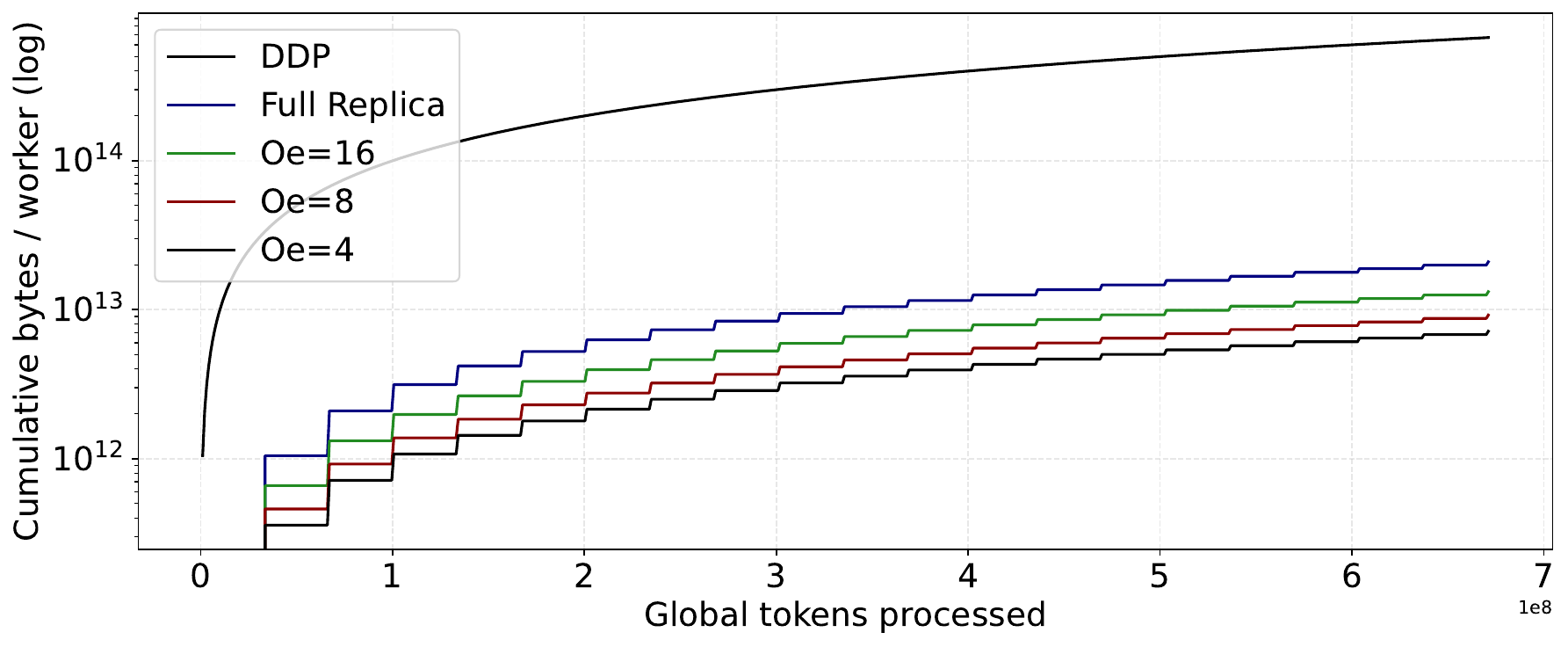}} \hfill
    \caption{\textbf{Analytical/modeled cumulative communication overhead versus global tokens.}
    Communication cost (log scale) for a large model ($d_{model}=8192$, 512 layers) on 32 workers using \diloco with random placement. \method achieves increasing bandwidth savings over \texttt{DDP} and Full Replica as expert overlap decreases from $O_e=16$ (green) to $O_e=4$ under the model assumptions.}
    \label{fig:comms_scaling_random}
\end{figure}

\vspace{0.3em}
\noindent\textbf{Communication Scalability.}\label{eval:theory:comms}
\Cref{fig:comms_scaling_random} shows modeled cumulative communication in which \method reduces bytes versus \texttt{DDP} and full-replica baselines. Unlike \texttt{DDP}'s continuous overhead, \method employs periodic, smaller bursts.

\begin{figure}[htb]
    \centering
        {\includegraphics[width=.8\columnwidth]{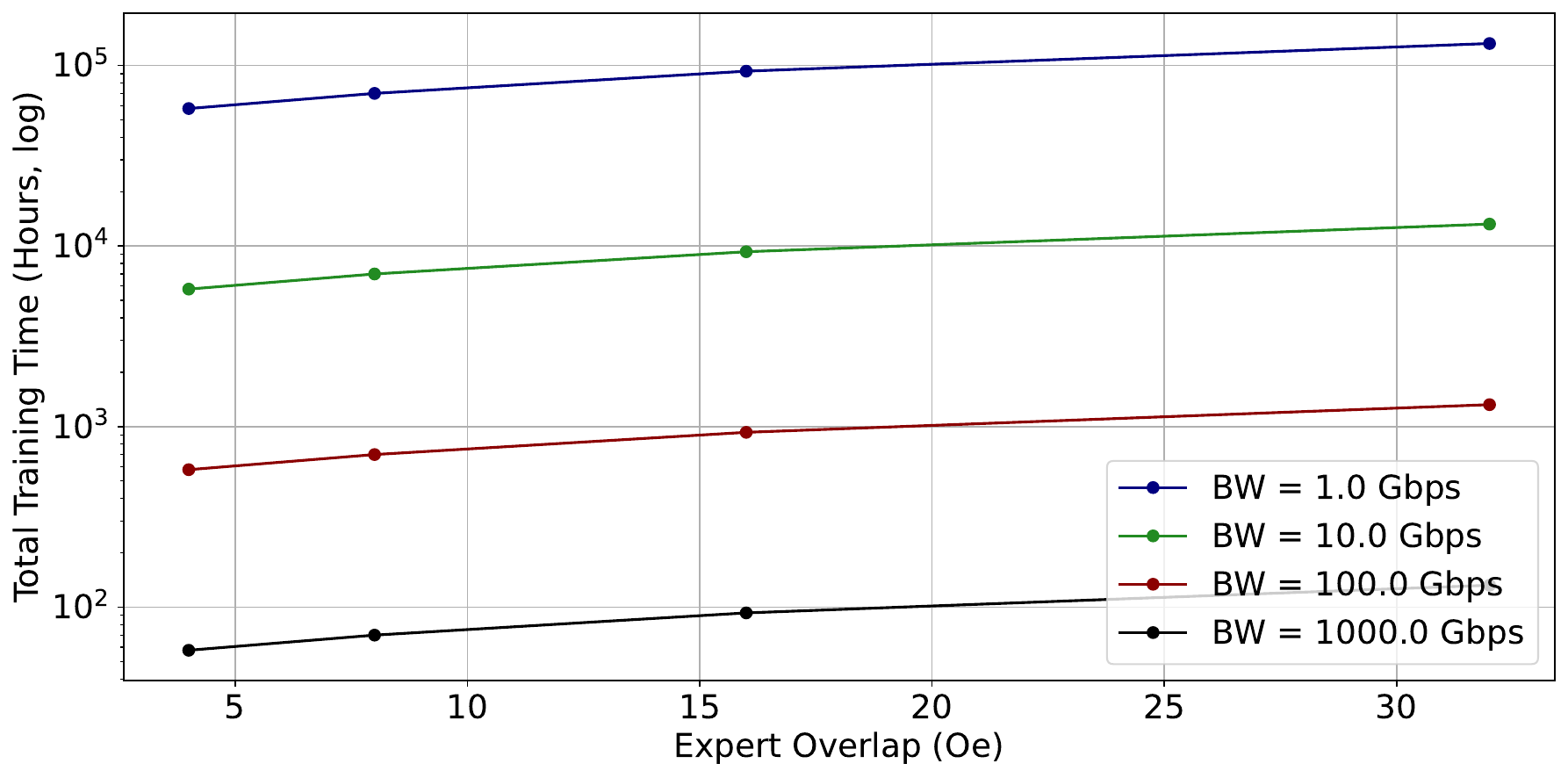}} \hfill
    \caption{\textbf{Analytical/modeled total training time versus expert overlap ($O_e$).}
    Modeled time for a model ($d_{model}=2048$, 64 layers) trained for 60B tokens on 32 H100s using \diloco (random). Reduced overlap accelerates training, especially under bandwidth constraints.}
    \label{fig:time_scaling}
\end{figure}

\vspace{+0.3em}
\noindent\textbf{Wall-clock Time Scalability.}\label{eval:theory:time}
\Cref{fig:time_scaling} demonstrates that minimizing communication via lower $O_e$ reduces modeled training time across the plotted bandwidths (1--1000 Gbps). Speedups are most critical in low-bandwidth regimes (e.g., 1 Gbps), where minimizing overlap is essential. In these modeled bandwidth regimes, \method provides a benefit over the full-replica variant. Since the gains provided by \method are multiplicative with communication-efficient methods, their relative magnitude holds across bandwidths under the cost-model assumptions.

\begin{figure}[htb]
    \centering
        {\includegraphics[width=.8\columnwidth]{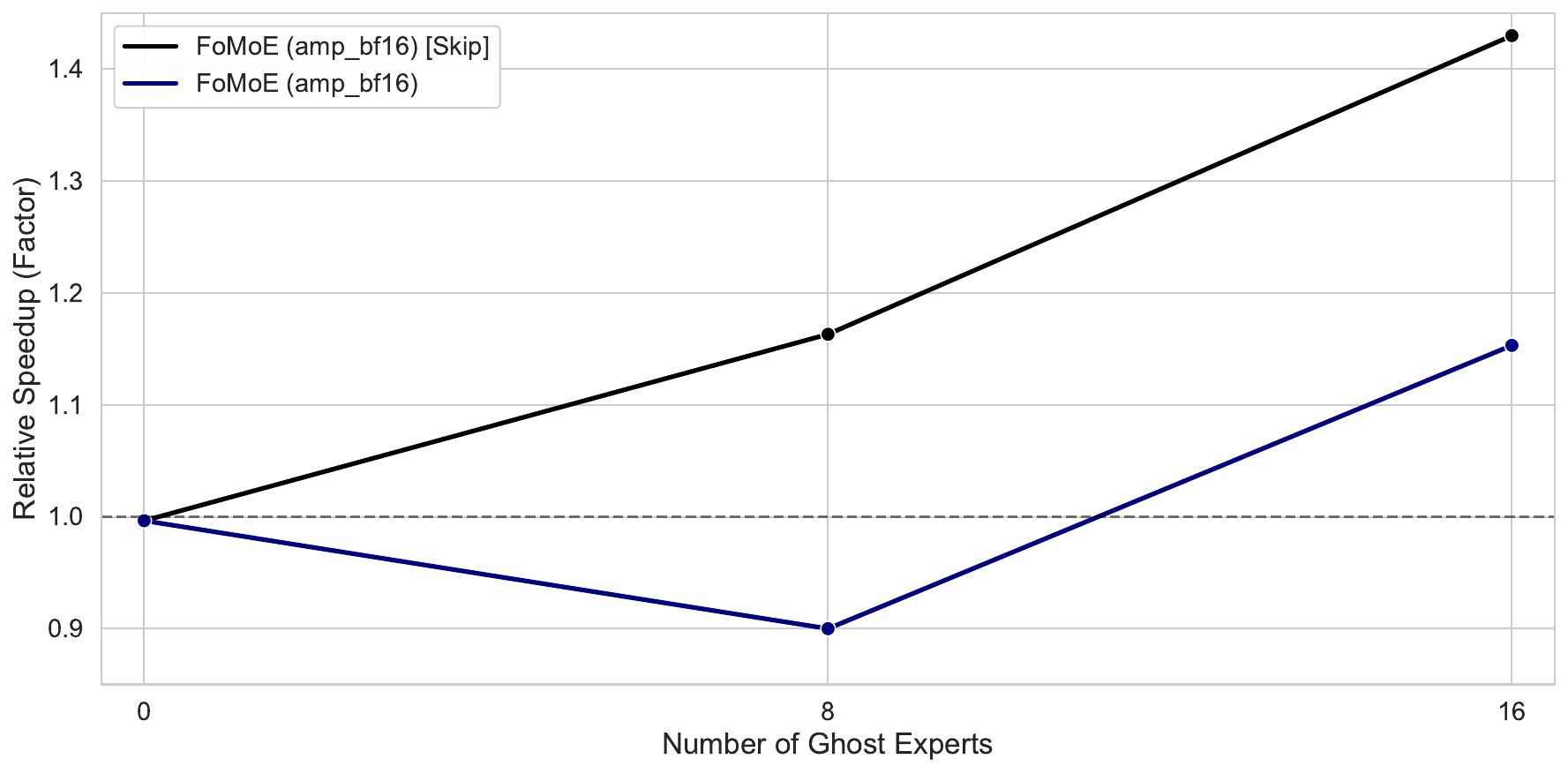}} \hfill
    \caption{\textbf{Measured speedup from token skipping.}
    Relative speedup versus the number of ghost experts for a 150M MoE model with $d_{\mathrm{model}}=512$ and expert size $128$ on 8 NVIDIA B200s. Skipping tokens achieves near-linear scaling, thereby validating the theoretical savings despite the lack of a custom kernel.}
    \label{fig:skip_tokens_throughput}
    \vspace{-0.1cm}
\end{figure}

\noindent\textbf{Measured Skip-Token Throughput.}
\Cref{fig:skip_tokens_throughput} confirms that the ``skip-token'' mechanism enables the linear throughput gains predicted in \cref{sec:scalability-costs} by bypassing computation for ghost experts. While reducing expert overlap generally improves throughput, skipping-token scales nearly linearly and outperforms the baseline.
The current implementation of the mechanism is limited by a lack of a dedicated kernel; however, we expect that the theoretical gains will be at least partially transferable to any optimized implementation.

\section{Limitations}\label{sec:limits}

Our work is limited in scale. We train proxy-size models and measure skip-token throughput on accelerator hardware; the XL/XXL and $100$B-scale results are projections from the communication, memory, \flop, and wall-clock models, not end-to-end pre-training runs.
Skip-token is also not equivalent to full-replica training or EP with remote activation dispatch: tokens routed to non-local experts are skipped rather than sent over the \wan. We evaluate its ML effect through validation perplexity, while downstream task performance remains future work.
The system setting is deliberately controlled. Our evaluation assumes homogeneous workers, static traffic shaping, synchronous rounds, and representative \wan bandwidth/latency. It does not yet cover time-varying jitter, asymmetric links, persistent stragglers, hierarchical \wan topologies, or heterogeneous GPU clusters. Since the present implementation is synchronous, a slow site can still delay the corresponding synchronization round.
Finally, the baseline and optimization scope are narrower than the full design space. We focus on \texttt{DDP} and full-replica low-communication methods such as \diloco/\photon-style training; a direct \wan-deployed EP baseline would be a useful stress test, although standard EP assumes high-bandwidth activation exchange and is therefore not the bottleneck targeted by \method. We also use standard low-frequency distributed optimization, adapted to partial expert replication, in order to isolate the payload-size effect. Future work should co-design optimizers, placement schedules, and rules for migrating optimizer state specifically for partial expert replication.

\section{Conclusion}

This paper presents \method, a framework that exploits \moe sparsity at the synchronization boundary in cross-datacenter training. Existing low-communication methods primarily reduce \emph{how often} sites synchronize; \method targets the complementary bottleneck of \emph{how much} model state crosses the \wan at each round. By partitioning expert layers, \method reduces communication overhead by up to $1.42\times$ over efficient baselines and $45.44\times$ over \texttt{DDP} in the studied regimes, while the ``skip-token'' mechanism delivers measured throughput speedups of up to $1.4\times$. The trained proxy regimes exhibit stable routing and convergence behavior, and the system model predicts that the same payload and memory reductions become increasingly important as sparse experts dominate larger configurations. Given these results, we argue that \moe sparsity should not stop at the datacenter boundary: when \wan bandwidth is the scarce resource, reducing the synchronization payload can be an effective communication-reduction mechanism alongside reducing synchronization frequency.

\section*{Acknowledgments}

This research was supported by the following entities: The Royal Academy of Engineering via DANTE (a RAEng Chair); the European Research Council, specifically the REDIAL project; SPRIND under the composite learning challenge; Google through a Google Academic Research Award; in addition to both IMEC and the Ministry of Education of Romania (through the Credit and Scholarship Agency).

\bibliographystyle{\paperbibstyle}
\bibliography{references}

\begin{thebibliography}{73}
\providecommand{\natexlab}[1]{#1}
\providecommand{\url}[1]{\texttt{#1}}
\expandafter\ifx\csname urlstyle\endcsname\relax
  \providecommand{\doi}[1]{doi: #1}\else
  \providecommand{\doi}{doi: \begingroup \urlstyle{rm}\Url}\fi

\bibitem[Abadi et~al.(2016)Abadi, Chu, Goodfellow, McMahan, Mironov, Talwar,
  and Zhang]{abadi2016dpsgd}
Abadi, M., Chu, A., Goodfellow, I.~J., McMahan, H.~B., Mironov, I., Talwar, K.,
  and Zhang, L.
\newblock Deep learning with differential privacy.
\newblock In \emph{{CCS}}, pp.\  308--318. {ACM}, 2016.

\bibitem[Alistarh et~al.(2017)Alistarh, Grubic, Li, Tomioka, and
  Vojnovic]{alistarh2017qsgd}
Alistarh, D., Grubic, D., Li, J., Tomioka, R., and Vojnovic, M.
\newblock {QSGD:} communication-efficient {SGD} via gradient quantization and
  encoding.
\newblock In \emph{{NIPS}}, pp.\  1709--1720, 2017.

\bibitem[Athlur et~al.(2022)Athlur, Saran, Sivathanu, Ramjee, and
  Kwatra]{athlur2022varuna}
Athlur, S., Saran, N., Sivathanu, M., Ramjee, R., and Kwatra, N.
\newblock Varuna: scalable, low-cost training of massive deep learning models.
\newblock In \emph{EuroSys}, pp.\  472--487. {ACM}, 2022.

\bibitem[Ben~Allal et~al.(2024)Ben~Allal, Lozhkov, Penedo, Wolf, and von
  Werra]{benallal2024smollmcorpus}
Ben~Allal, L., Lozhkov, A., Penedo, G., Wolf, T., and von Werra, L.
\newblock {SmolLM-Corpus}, July 2024.
\newblock URL
  \url{https://huggingface.co/datasets/HuggingFaceTB/smollm-corpus}.

\bibitem[Bonawitz et~al.(2017)Bonawitz, Ivanov, Kreuter, Marcedone, McMahan,
  Patel, Ramage, Segal, and Seth]{bonawitz2017secagg}
Bonawitz, K.~A., Ivanov, V., Kreuter, B., Marcedone, A., McMahan, H.~B., Patel,
  S., Ramage, D., Segal, A., and Seth, K.
\newblock Practical secure aggregation for privacy-preserving machine learning.
\newblock In \emph{{CCS}}, pp.\  1175--1191. {ACM}, 2017.

\bibitem[Cagnasso et~al.(2026)Cagnasso, Belilovsky, and
  Oyallon]{cagnasso2026gasloc}
Cagnasso, P., Belilovsky, E., and Oyallon, E.
\newblock Unifying local communications and local updates for {LLM}
  pretraining.
\newblock \emph{CoRR}, abs/2606.11081, 2026.
\newblock \doi{10.48550/arXiv.2606.11081}.
\newblock URL \url{https://arxiv.org/abs/2606.11081}.

\bibitem[Chen et~al.(2016)Chen, Xu, Zhang, and Guestrin]{chen2016checkpointing}
Chen, T., Xu, B., Zhang, C., and Guestrin, C.
\newblock Training deep nets with sublinear memory cost, 2016.
\newblock URL \url{https://arxiv.org/abs/1604.06174}.

\bibitem[Cheng \& Glasgow(2025)Cheng and Glasgow]{cheng2025localadam}
Cheng, Z. and Glasgow, M.
\newblock Convergence of distributed adaptive optimization with local updates.
\newblock In \emph{{ICLR}}. OpenReview.net, 2025.

\bibitem[Comanici et~al.(2025)Comanici, Bieber, Schaekermann, Pasupat,
  Sachdeva, et~al.]{comanici2025gemini25pushingfrontier}
Comanici, G., Bieber, E., Schaekermann, M., Pasupat, I., Sachdeva, N., et~al.
\newblock Gemini 2.5: Pushing the frontier with advanced reasoning,
  multimodality, long context, and next generation agentic capabilities.
\newblock \emph{CoRR}, abs/2507.06261, 2025.
\newblock \doi{10.48550/arXiv.2507.06261}.
\newblock URL \url{https://arxiv.org/abs/2507.06261}.

\bibitem[Csord{\'{a}}s et~al.(2023)Csord{\'{a}}s, Irie, and
  Schmidhuber]{csordas2023sigma}
Csord{\'{a}}s, R., Irie, K., and Schmidhuber, J.
\newblock Approximating two-layer feedforward networks for efficient
  transformers.
\newblock In \emph{{EMNLP} (Findings)}, pp.\  674--692. Association for
  Computational Linguistics, 2023.

\bibitem[{DeepSeek-AI}(2024)]{deepseekv3}
{DeepSeek-AI}.
\newblock Deepseek-v3 technical report.
\newblock \emph{CoRR}, abs/2412.19437, 2024.
\newblock URL \url{https://arxiv.org/abs/2412.19437}.

\bibitem[DeepSeek{-}AI et~al.(2024)DeepSeek{-}AI, Liu, Feng, Wang, Wang, Liu,
  Zhao, Deng, Ruan, Dai, Guo, Yang, Chen, Ji, Li, Lin, Luo, Hao, Chen, Li,
  Zhang, Xu, Yang, Zhang, Ding, Xin, Gao, Li, Qu, Cai, Liang, Guo, Ni, Li,
  Chen, Yuan, Qiu, Song, Dong, Gao, Guan, Wang, Zhang, Xu, Xia, Zhao, Zhang,
  Li, Wang, Zhang, Zhang, Tang, Li, Tian, Huang, Wang, Zhang, Zhu, Chen, Du,
  Chen, Jin, Ge, Pan, Xu, Chen, Li, Lu, Zhou, Chen, Wu, Ye, Ma, Wang, Zhou, Yu,
  Zhou, Zheng, Wang, Pei, Yuan, Sun, Xiao, Zeng, An, Liu, Liang, Gao, Zhang,
  Li, Jin, Wang, Bi, Liu, Wang, Shen, Chen, Chen, Nie, Sun, Wang, and
  et~al.]{liu2024deepseekv2}
DeepSeek{-}AI, Liu, A., Feng, B., Wang, B., Wang, B., Liu, B., Zhao, C., Deng,
  C., Ruan, C., Dai, D., Guo, D., Yang, D., Chen, D., Ji, D., Li, E., Lin, F.,
  Luo, F., Hao, G., Chen, G., Li, G., Zhang, H., Xu, H., Yang, H., Zhang, H.,
  Ding, H., Xin, H., Gao, H., Li, H., Qu, H., Cai, J.~L., Liang, J., Guo, J.,
  Ni, J., Li, J., Chen, J., Yuan, J., Qiu, J., Song, J., Dong, K., Gao, K.,
  Guan, K., Wang, L., Zhang, L., Xu, L., Xia, L., Zhao, L., Zhang, L., Li, M.,
  Wang, M., Zhang, M., Zhang, M., Tang, M., Li, M., Tian, N., Huang, P., Wang,
  P., Zhang, P., Zhu, Q., Chen, Q., Du, Q., Chen, R.~J., Jin, R.~L., Ge, R.,
  Pan, R., Xu, R., Chen, R., Li, S.~S., Lu, S., Zhou, S., Chen, S., Wu, S., Ye,
  S., Ma, S., Wang, S., Zhou, S., Yu, S., Zhou, S., Zheng, S., Wang, T., Pei,
  T., Yuan, T., Sun, T., Xiao, W.~L., Zeng, W., An, W., Liu, W., Liang, W.,
  Gao, W., Zhang, W., Li, X.~Q., Jin, X., Wang, X., Bi, X., Liu, X., Wang, X.,
  Shen, X., Chen, X., Chen, X., Nie, X., Sun, X., Wang, Z., and et~al.
\newblock Deepseek-v2: {A} strong, economical, and efficient mixture-of-experts
  language model.
\newblock \emph{CoRR}, abs/2405.04434, 2024.

\bibitem[Dey et~al.(2025)Dey, Zhang, Noci, Li, Bordelon, Bergsma, Pehlevan,
  Hanin, and Hestness]{dey2025completep}
Dey, N., Zhang, B.~C., Noci, L., Li, M.~B., Bordelon, B., Bergsma, S.,
  Pehlevan, C., Hanin, B., and Hestness, J.
\newblock Don't be lazy: Completep enables compute-efficient deep transformers.
\newblock \emph{CoRR}, abs/2505.01618, 2025.

\bibitem[Douillard et~al.(2023)Douillard, Feng, Rusu, Chhaparia, Donchev,
  Kuncoro, Ranzato, Szlam, and Shen]{douillard2023diloco}
Douillard, A., Feng, Q., Rusu, A.~A., Chhaparia, R., Donchev, Y., Kuncoro, A.,
  Ranzato, M., Szlam, A., and Shen, J.
\newblock Diloco: Distributed low-communication training of language models.
\newblock \emph{CoRR}, abs/2311.08105, 2023.
\newblock \doi{10.48550/arXiv.2311.08105}.
\newblock URL \url{https://arxiv.org/abs/2311.08105}.

\bibitem[Douillard et~al.(2024)Douillard, Feng, Rusu, Kuncoro, Donchev,
  Chhaparia, Gog, Ranzato, Shen, and Szlam]{douillard2024dipaco}
Douillard, A., Feng, Q., Rusu, A.~A., Kuncoro, A., Donchev, Y., Chhaparia, R.,
  Gog, I., Ranzato, M., Shen, J., and Szlam, A.
\newblock Dipaco: Distributed path composition.
\newblock \emph{CoRR}, abs/2403.10616, 2024.

\bibitem[Douillard et~al.(2025)Douillard, Donchev, Rush, Kale, Charles,
  Garrett, Teston, Lacey, McIlroy, Shen, Ram{\'{e}}, Szlam, Ranzato, and
  Barham]{douillard2025streaming}
Douillard, A., Donchev, Y., Rush, K., Kale, S., Charles, Z., Garrett, Z.,
  Teston, G., Lacey, D., McIlroy, R., Shen, J., Ram{\'{e}}, A., Szlam, A.,
  Ranzato, M., and Barham, P.
\newblock Streaming diloco with overlapping communication: Towards a
  distributed free lunch.
\newblock \emph{CoRR}, abs/2501.18512, 2025.

\bibitem[Du et~al.(2022)Du, Huang, Dai, Tong, Lepikhin, Xu, Krikun, Zhou, Yu,
  Firat, Zoph, Fedus, Bosma, Zhou, Wang, Wang, Webster, Pellat, Robinson,
  Meier{-}Hellstern, Duke, Dixon, Zhang, Le, Wu, Chen, and Cui]{du2022glam}
Du, N., Huang, Y., Dai, A.~M., Tong, S., Lepikhin, D., Xu, Y., Krikun, M.,
  Zhou, Y., Yu, A.~W., Firat, O., Zoph, B., Fedus, L., Bosma, M.~P., Zhou, Z.,
  Wang, T., Wang, Y.~E., Webster, K., Pellat, M., Robinson, K.,
  Meier{-}Hellstern, K.~S., Duke, T., Dixon, L., Zhang, K., Le, Q.~V., Wu, Y.,
  Chen, Z., and Cui, C.
\newblock Glam: Efficient scaling of language models with mixture-of-experts.
\newblock In \emph{{ICML}}, volume 162 of \emph{Proceedings of Machine Learning
  Research}, pp.\  5547--5569. {PMLR}, 2022.

\bibitem[Fedus et~al.(2022)Fedus, Zoph, and Shazeer]{fedus2021switch}
Fedus, W., Zoph, B., and Shazeer, N.
\newblock Switch transformers: Scaling to trillion parameter models with simple
  and efficient sparsity.
\newblock \emph{J. Mach. Learn. Res.}, 23:\penalty0 120:1--120:39, 2022.

\bibitem[Gale et~al.(2023)Gale, Narayanan, Young, and
  Zaharia]{gale2022megablocks}
Gale, T., Narayanan, D., Young, C., and Zaharia, M.
\newblock Megablocks: Efficient sparse training with mixture-of-experts.
\newblock In \emph{MLSys}. mlsys.org, 2023.

\bibitem[Guo et~al.(2021)Guo, Mei, Xiao, and Wu]{guo2021pfl}
Guo, B., Mei, Y., Xiao, D., and Wu, W.
\newblock {PFL-MoE}: Personalized federated learning based on mixture of
  experts.
\newblock In \emph{Web and Big Data}, volume 12858 of \emph{Lecture Notes in
  Computer Science}, pp.\  480--486. Springer, 2021.
\newblock \doi{10.1007/978-3-030-85896-4_37}.
\newblock URL \url{https://doi.org/10.1007/978-3-030-85896-4_37}.

\bibitem[Guthrie(2025)]{Guthrie_2025}
Guthrie, S.
\newblock Inside the world’s most powerful ai datacenter, Sept 2025.
\newblock URL
  \url{https://blogs.microsoft.com/blog/2025/09/18/inside-the-worlds-most-powerful-ai-datacenter/}.

\bibitem[He et~al.(2022)He, Zhai, Antunes, Wang, Luo, Shi, and
  Li]{he2022fastermoe}
He, J., Zhai, J., Antunes, T., Wang, H., Luo, F., Shi, S., and Li, Q.
\newblock Fastermoe: modeling and optimizing training of large-scale dynamic
  pre-trained models.
\newblock In \emph{PPoPP}, pp.\  120--134. {ACM}, 2022.

\bibitem[Hoffmann et~al.(2022)Hoffmann, Borgeaud, Mensch, Buchatskaya, Cai,
  Rutherford, de~Las~Casas, Hendricks, Welbl, Clark, Hennigan, Noland,
  Millican, van~den Driessche, Damoc, Guy, Osindero, Simonyan, Elsen, Rae,
  Vinyals, and Sifre]{hoffmann2022chinchilla}
Hoffmann, J., Borgeaud, S., Mensch, A., Buchatskaya, E., Cai, T., Rutherford,
  E., de~Las~Casas, D., Hendricks, L.~A., Welbl, J., Clark, A., Hennigan, T.,
  Noland, E., Millican, K., van~den Driessche, G., Damoc, B., Guy, A.,
  Osindero, S., Simonyan, K., Elsen, E., Rae, J.~W., Vinyals, O., and Sifre, L.
\newblock Training compute-optimal large language models.
\newblock \emph{CoRR}, abs/2203.15556, 2022.

\bibitem[Huang et~al.(2019)Huang, Cheng, Bapna, Firat, Chen, Chen, Lee, Ngiam,
  Le, Wu, and Chen]{huang2019gpipe}
Huang, Y., Cheng, Y., Bapna, A., Firat, O., Chen, D., Chen, M.~X., Lee, H.,
  Ngiam, J., Le, Q.~V., Wu, Y., and Chen, Z.
\newblock Gpipe: Efficient training of giant neural networks using pipeline
  parallelism.
\newblock In \emph{NeurIPS}, pp.\  103--112, 2019.

\bibitem[Hwang et~al.(2023)Hwang, Cui, Xiong, Yang, Liu, Hu, Wang, Salas, Jose,
  Ram, Chau, Cheng, Yang, Yang, and Xiong]{hwang2022tutel}
Hwang, C., Cui, W., Xiong, Y., Yang, Z., Liu, Z., Hu, H., Wang, Z., Salas, R.,
  Jose, J., Ram, P., Chau, J., Cheng, P., Yang, F., Yang, M., and Xiong, Y.
\newblock Tutel: Adaptive mixture-of-experts at scale.
\newblock In \emph{MLSys}. mlsys.org, 2023.

\bibitem[Iacob et~al.(2025{\natexlab{a}})Iacob, Sani, Kurmanji, Shen, Qiu, Cai,
  Gao, and Lane]{iacob2024dept}
Iacob, A., Sani, L., Kurmanji, M., Shen, W.~F., Qiu, X., Cai, D., Gao, Y., and
  Lane, N.~D.
\newblock {DEPT:} decoupled embeddings for pre-training language models.
\newblock In \emph{{ICLR}}. OpenReview.net, 2025{\natexlab{a}}.

\bibitem[Iacob et~al.(2025{\natexlab{b}})Iacob, Sani, Safaryan, Giampouras,
  Horv{\'{a}}th, Jovanovic, Kurmanji, Aleksandrov, Shen, Qiu, and
  Lane]{iacob2025desloc}
Iacob, A., Sani, L., Safaryan, M., Giampouras, P., Horv{\'{a}}th, S.,
  Jovanovic, A., Kurmanji, M., Aleksandrov, P., Shen, W.~F., Qiu, X., and Lane,
  N.~D.
\newblock {DES-LOC:} desynced low communication adaptive optimizers for
  training foundation models.
\newblock \emph{CoRR}, abs/2505.22549, 2025{\natexlab{b}}.

\bibitem[Jaghouar et~al.(2024)Jaghouar, Ong, and
  Hagemann]{jaghouar2024opendiloco}
Jaghouar, S., Ong, J.~M., and Hagemann, J.
\newblock Opendiloco: An open-source framework for globally distributed
  low-communication training.
\newblock \emph{CoRR}, abs/2407.07852, 2024.

\bibitem[Kairouz et~al.(2021)Kairouz, McMahan, Avent, Bellet, Bennis, Bhagoji,
  Bonawitz, Charles, Cormode, Cummings, D'Oliveira, Eichner, Rouayheb, Evans,
  Gardner, Garrett, Gasc{\'{o}}n, Ghazi, Gibbons, Gruteser, Harchaoui, He, He,
  Huo, Hutchinson, Hsu, Jaggi, Javidi, Joshi, Khodak, Kone{\v{c}}n{\'y},
  Korolova, Koushanfar, Koyejo, Lepoint, Liu, Mittal, Mohri, Nock,
  {\"{O}}zg{\"{u}}r, Pagh, Qi, Ramage, Raskar, Raykova, Song, Song, Stich, Sun,
  Suresh, Tram{\`{e}}r, Vepakomma, Wang, Xiong, Xu, Yang, Yu, Yu, and
  Zhao]{kairouz2021advances}
Kairouz, P., McMahan, H.~B., Avent, B., Bellet, A., Bennis, M., Bhagoji, A.~N.,
  Bonawitz, K.~A., Charles, Z., Cormode, G., Cummings, R., D'Oliveira, R.
  G.~L., Eichner, H., Rouayheb, S.~E., Evans, D., Gardner, J., Garrett, Z.,
  Gasc{\'{o}}n, A., Ghazi, B., Gibbons, P.~B., Gruteser, M., Harchaoui, Z., He,
  C., He, L., Huo, Z., Hutchinson, B., Hsu, J., Jaggi, M., Javidi, T., Joshi,
  G., Khodak, M., Kone{\v{c}}n{\'y}, J., Korolova, A., Koushanfar, F., Koyejo,
  S., Lepoint, T., Liu, Y., Mittal, P., Mohri, M., Nock, R., {\"{O}}zg{\"{u}}r,
  A., Pagh, R., Qi, H., Ramage, D., Raskar, R., Raykova, M., Song, D., Song,
  W., Stich, S.~U., Sun, Z., Suresh, A.~T., Tram{\`{e}}r, F., Vepakomma, P.,
  Wang, J., Xiong, L., Xu, Z., Yang, Q., Yu, F.~X., Yu, H., and Zhao, S.
\newblock Advances and open problems in federated learning.
\newblock \emph{Found. Trends Mach. Learn.}, 14\penalty0 (1-2):\penalty0
  1--210, 2021.

\bibitem[Kaplan et~al.(2020)Kaplan, McCandlish, Henighan, Brown, Chess, Child,
  Gray, Radford, Wu, and Amodei]{kaplan2020scalinglaws}
Kaplan, J., McCandlish, S., Henighan, T., Brown, T.~B., Chess, B., Child, R.,
  Gray, S., Radford, A., Wu, J., and Amodei, D.
\newblock Scaling laws for neural language models.
\newblock \emph{CoRR}, abs/2001.08361, 2020.

\bibitem[Khaled et~al.(2025)Khaled, Kale, Douillard, Jin, Fergus, and
  Zaheer]{khaled2025outer}
Khaled, A., Kale, S., Douillard, A., Jin, C., Fergus, R., and Zaheer, M.
\newblock Understanding outer optimizers in local {SGD:} learning rates,
  momentum, and acceleration.
\newblock \emph{CoRR}, abs/2509.10439, 2025.

\bibitem[Kim et~al.(2025)Kim, Lee, Park, Oh, Kim, Yoo, Shin, Han, Shin, and
  Yoo]{PeriLN}
Kim, J., Lee, B., Park, C., Oh, Y., Kim, B., Yoo, T., Shin, S., Han, D., Shin,
  J., and Yoo, K.~M.
\newblock Peri-ln: Revisiting normalization layer in the transformer
  architecture.
\newblock In \emph{{ICML}}. OpenReview.net, 2025.

\bibitem[Lepikhin et~al.(2021)Lepikhin, Lee, Xu, Chen, Firat, Huang, Krikun,
  Shazeer, and Chen]{lepikhin2021gshard}
Lepikhin, D., Lee, H., Xu, Y., Chen, D., Firat, O., Huang, Y., Krikun, M.,
  Shazeer, N., and Chen, Z.
\newblock Gshard: Scaling giant models with conditional computation and
  automatic sharding.
\newblock In \emph{{ICLR}}. OpenReview.net, 2021.

\bibitem[Lewis et~al.(2021)Lewis, Bhosale, Dettmers, Goyal, and
  Zettlemoyer]{lewis2021base}
Lewis, M., Bhosale, S., Dettmers, T., Goyal, N., and Zettlemoyer, L.
\newblock {BASE} layers: Simplifying training of large, sparse models.
\newblock In \emph{{ICML}}, volume 139 of \emph{Proceedings of Machine Learning
  Research}, pp.\  6265--6274. {PMLR}, 2021.

\bibitem[Li et~al.(2020{\natexlab{a}})Li, Zhao, Varma, Salpekar, Noordhuis, Li,
  Paszke, Smith, Vaughan, Damania, et~al.]{li2020pytorch}
Li, S., Zhao, Y., Varma, R., Salpekar, O., Noordhuis, P., Li, T., Paszke, A.,
  Smith, J., Vaughan, B., Damania, P., et~al.
\newblock Pytorch distributed: Experiences on accelerating data parallel
  training.
\newblock \emph{arXiv preprint arXiv:2006.15704}, 2020{\natexlab{a}}.

\bibitem[Li et~al.(2020{\natexlab{b}})Li, Sahu, Zaheer, Sanjabi, Talwalkar, and
  Smith]{li2020fedprox}
Li, T., Sahu, A.~K., Zaheer, M., Sanjabi, M., Talwalkar, A., and Smith, V.
\newblock Federated optimization in heterogeneous networks.
\newblock In \emph{MLSys}. mlsys.org, 2020{\natexlab{b}}.

\bibitem[Loshchilov \& Hutter(2019)Loshchilov and Hutter]{loshchilov2019adamw}
Loshchilov, I. and Hutter, F.
\newblock Decoupled weight decay regularization.
\newblock In \emph{{ICLR} (Poster)}. OpenReview.net, 2019.

\bibitem[Lu et~al.(2024)Lu, Liu, Xu, Zhou, Huang, Zhang, Yan, and
  Li]{lu2024notall}
Lu, X., Liu, Q., Xu, Y., Zhou, A., Huang, S., Zhang, B., Yan, J., and Li, H.
\newblock Not all experts are equal: Efficient expert pruning and skipping for
  mixture-of-experts large language models.
\newblock In \emph{Proceedings of the 62nd Annual Meeting of the Association
  for Computational Linguistics (Volume 1: Long Papers)}, pp.\  6159--6172.
  Association for Computational Linguistics, 2024.
\newblock \doi{10.18653/v1/2024.acl-long.334}.
\newblock URL \url{https://aclanthology.org/2024.acl-long.334/}.

\bibitem[Ludziejewski et~al.(2024)Ludziejewski, Krajewski, Adamczewski,
  Pi{\'{o}}ro, Krutul, Antoniak, Ciebiera, Kr{\'{o}}l, Odrzyg{\'{o}}zdz,
  Sankowski, Cygan, and Jaszczur]{krajewski2024fgmoe}
Ludziejewski, J., Krajewski, J., Adamczewski, K., Pi{\'{o}}ro, M., Krutul, M.,
  Antoniak, S., Ciebiera, K., Kr{\'{o}}l, K., Odrzyg{\'{o}}zdz, T., Sankowski,
  P., Cygan, M., and Jaszczur, S.
\newblock Scaling laws for fine-grained mixture of experts.
\newblock In \emph{{ICML}}. OpenReview.net, 2024.

\bibitem[Malasnicki et~al.(2025)Malasnicki, Ciebiera, Borun, Pi{\'{o}}ro,
  Ludziejewski, Stefaniak, Krutul, Jaszczur, Cygan, Adamczewski, and
  Krajewski]{malasnicki2025muparam}
Malasnicki, J., Ciebiera, K., Borun, M., Pi{\'{o}}ro, M., Ludziejewski, J.,
  Stefaniak, M., Krutul, M., Jaszczur, S., Cygan, M., Adamczewski, K., and
  Krajewski, J.
\newblock {\(\mu\)}-parametrization for mixture of experts.
\newblock \emph{CoRR}, abs/2508.09752, 2025.

\bibitem[McMahan et~al.(2017)McMahan, Moore, Ramage, Hampson, and
  y~Arcas]{mcmahan2017fedavg}
McMahan, B., Moore, E., Ramage, D., Hampson, S., and y~Arcas, B.~A.
\newblock Communication-efficient learning of deep networks from decentralized
  data.
\newblock In \emph{{AISTATS}}, volume~54 of \emph{Proceedings of Machine
  Learning Research}, pp.\  1273--1282. {PMLR}, 2017.

\bibitem[Mei et~al.(2024)Mei, Cai, Zhou, Wang, and Xu]{mei2024fedmoe}
Mei, H., Cai, D., Zhou, A., Wang, S., and Xu, M.
\newblock Fedmoe: Personalized federated learning via heterogeneous mixture of
  experts.
\newblock \emph{arXiv preprint arXiv:2408.11304}, 2024.

\bibitem[Micikevicius et~al.(2018)Micikevicius, Narang, Alben, Diamos, Elsen,
  Garc{\'{\i}}a, Ginsburg, Houston, Kuchaiev, Venkatesh, and
  Wu]{micikevicius2018mixed}
Micikevicius, P., Narang, S., Alben, J., Diamos, G.~F., Elsen, E.,
  Garc{\'{\i}}a, D., Ginsburg, B., Houston, M., Kuchaiev, O., Venkatesh, G.,
  and Wu, H.
\newblock Mixed precision training.
\newblock In \emph{{ICLR} (Poster)}. OpenReview.net, 2018.

\bibitem[Narayanan et~al.(2019)Narayanan, Harlap, Phanishayee, Seshadri,
  Devanur, Ganger, Gibbons, and Zaharia]{narayanan2019pipedream}
Narayanan, D., Harlap, A., Phanishayee, A., Seshadri, V., Devanur, N.~R.,
  Ganger, G.~R., Gibbons, P.~B., and Zaharia, M.
\newblock Pipedream: generalized pipeline parallelism for {DNN} training.
\newblock In \emph{{SOSP}}, pp.\  1--15. {ACM}, 2019.

\bibitem[OpenAI(2025)]{gptoss}
OpenAI.
\newblock gpt-oss-120b {\&} gpt-oss-20b model card.
\newblock \emph{CoRR}, abs/2508.10925, 2025.

\bibitem[Palak et~al.(2024)Palak, Reddy, Kataria, Gandhi, Tandon,
  Bhattacherjee, and Padmanabhan]{palak2024geotraining}
Palak, Reddy, T.~R., Kataria, B., Gandhi, R., Tandon, K., Bhattacherjee, D.,
  and Padmanabhan, V.~N.
\newblock Improving training time and {GPU} utilization in geo-distributed
  language model training.
\newblock \emph{CoRR}, abs/2411.14458, 2024.
\newblock \doi{10.48550/arXiv.2411.14458}.
\newblock URL \url{https://arxiv.org/abs/2411.14458}.

\bibitem[Patarasuk \& Yuan(2009)Patarasuk and Yuan]{patarasuk2009allreduce}
Patarasuk, P. and Yuan, X.
\newblock Bandwidth optimal all-reduce algorithms for clusters of workstations.
\newblock \emph{J. Parallel Distributed Comput.}, 69\penalty0 (2):\penalty0
  117--124, 2009.

\bibitem[Qi et~al.(2025)Qi, Zhu, Li, Wu, Wu, He, Gao, Zeng, and
  Heinrich]{qi2025dilocox}
Qi, J., Zhu, W., Li, L., Wu, M., Wu, Y., He, W., Gao, X., Zeng, J., and
  Heinrich, M.
\newblock Dilocox: {A} low-communication large-scale training framework for
  decentralized cluster.
\newblock \emph{CoRR}, abs/2506.21263, 2025.

\bibitem[Rabenseifner(2004)]{rabenseifner2004reduce}
Rabenseifner, R.
\newblock Optimization of collective reduction operations.
\newblock In \emph{International Conference on Computational Science}, volume
  3036 of \emph{Lecture Notes in Computer Science}, pp.\  1--9. Springer, 2004.

\bibitem[Raffel et~al.(2020)Raffel, Shazeer, Roberts, Lee, Narang, Matena,
  Zhou, Li, and Liu]{raffel2023exploringlimitstransferlearning}
Raffel, C., Shazeer, N., Roberts, A., Lee, K., Narang, S., Matena, M., Zhou,
  Y., Li, W., and Liu, P.~J.
\newblock Exploring the limits of transfer learning with a unified text-to-text
  transformer.
\newblock \emph{Journal of Machine Learning Research}, 21\penalty0
  (140):\penalty0 1--67, 2020.
\newblock URL \url{https://jmlr.org/papers/v21/20-074.html}.

\bibitem[Rajbhandari et~al.(2020)Rajbhandari, Rasley, Ruwase, and
  He]{rajbhandari2020zero}
Rajbhandari, S., Rasley, J., Ruwase, O., and He, Y.
\newblock Zero: memory optimizations toward training trillion parameter models.
\newblock In \emph{{SC}}, pp.\ ~20. {IEEE/ACM}, 2020.

\bibitem[Rajbhandari et~al.(2022)Rajbhandari, Li, Yao, Zhang, Aminabadi, Awan,
  Rasley, and He]{rajbhandari2022deepspeedmoe}
Rajbhandari, S., Li, C., Yao, Z., Zhang, M., Aminabadi, R.~Y., Awan, A.~A.,
  Rasley, J., and He, Y.
\newblock Deepspeed-moe: Advancing mixture-of-experts inference and training to
  power next-generation {AI} scale.
\newblock In \emph{{ICML}}, volume 162 of \emph{Proceedings of Machine Learning
  Research}, pp.\  18332--18346. {PMLR}, 2022.

\bibitem[Reddi et~al.(2021)Reddi, Charles, Zaheer, Garrett, Rush,
  Kone{\v{c}}n{\'y}, Kumar, and McMahan]{reddi2020fedopt}
Reddi, S.~J., Charles, Z., Zaheer, M., Garrett, Z., Rush, K.,
  Kone{\v{c}}n{\'y}, J., Kumar, S., and McMahan, H.~B.
\newblock Adaptive federated optimization.
\newblock In \emph{{ICLR}}. OpenReview.net, 2021.

\bibitem[Reisser et~al.(2021)Reisser, Louizos, Gavves, and
  Welling]{reisser2021fedmix}
Reisser, M., Louizos, C., Gavves, E., and Welling, M.
\newblock Federated mixture of experts.
\newblock \emph{CoRR}, abs/2107.06724, 2021.

\bibitem[Riquelme et~al.(2021)Riquelme, Puigcerver, Mustafa, Neumann, Jenatton,
  Pinto, Keysers, and Houlsby]{riquelme2021vmoe}
Riquelme, C., Puigcerver, J., Mustafa, B., Neumann, M., Jenatton, R., Pinto,
  A.~S., Keysers, D., and Houlsby, N.
\newblock Scaling vision with sparse mixture of experts.
\newblock In \emph{NeurIPS}, pp.\  8583--8595, 2021.

\bibitem[Roller et~al.(2021)Roller, Sukhbaatar, Szlam, and
  Weston]{roller2021hash}
Roller, S., Sukhbaatar, S., Szlam, A., and Weston, J.
\newblock Hash layers for large sparse models.
\newblock In \emph{NeurIPS}, pp.\  17555--17566, 2021.

\bibitem[Ryabinin \& Gusev(2020)Ryabinin and Gusev]{ryabinin2020towards}
Ryabinin, M. and Gusev, A.
\newblock Towards crowdsourced training of large neural networks using
  decentralized mixture-of-experts.
\newblock \emph{Advances in Neural Information Processing Systems},
  33:\penalty0 3659--3672, 2020.
\newblock \doi{10.48550/arXiv.2002.04013}.
\newblock URL \url{https://arxiv.org/abs/2002.04013}.

\bibitem[Ryabinin et~al.(2023)Ryabinin, Dettmers, Diskin, and
  Borzunov]{ryabinin2023swarm}
Ryabinin, M., Dettmers, T., Diskin, M., and Borzunov, A.
\newblock {SWARM} parallelism: Training large models can be surprisingly
  communication-efficient.
\newblock In \emph{Proceedings of the 40th International Conference on Machine
  Learning}, volume 202 of \emph{Proceedings of Machine Learning Research},
  pp.\  29416--29440. PMLR, 2023.
\newblock URL \url{https://proceedings.mlr.press/v202/ryabinin23a.html}.

\bibitem[Sani et~al.(2025)Sani, Iacob, Cao, Lee, Marino, Gao, Zhao, Cai, Li,
  Qiu, and Lane]{sani2024photon}
Sani, L., Iacob, A., Cao, Z., Lee, R., Marino, B., Gao, Y., Zhao, W., Cai, D.,
  Li, Z., Qiu, X., and Lane, N.~D.
\newblock Photon: Federated {LLM} pre-training.
\newblock In \emph{Eighth Conference on Machine Learning and Systems}, 2025.
\newblock URL \url{https://openreview.net/forum?id=AQgYcfg5EI}.

\bibitem[Sarfi et~al.(2025)Sarfi, Th{\'{e}}rien, Lidin, and
  Belilovsky]{sarfi2025sparseloco}
Sarfi, A., Th{\'{e}}rien, B., Lidin, J., and Belilovsky, E.
\newblock Communication efficient {LLM} pre-training with {SparseLoCo}.
\newblock \emph{CoRR}, abs/2508.15706, 2025.
\newblock \doi{10.48550/arXiv.2508.15706}.
\newblock URL \url{https://arxiv.org/abs/2508.15706}.

\bibitem[Sevilla \& Troynikov(2025)Sevilla and
  Troynikov]{epoch2025coulddecentralizedtrainingsolveaispowerproblem}
Sevilla, J. and Troynikov, A.
\newblock Could decentralized training solve ai’s power problem?, 2025.
\newblock URL
  \url{https://epoch.ai/blog/could-decentralized-training-solve-ais-power-problem}.
\newblock Accessed: 2025-12-11.

\bibitem[Shazeer et~al.(2017)Shazeer, Mirhoseini, Maziarz, Davis, Le, Hinton,
  and Dean]{shazeer2017moe}
Shazeer, N., Mirhoseini, A., Maziarz, K., Davis, A., Le, Q.~V., Hinton, G.~E.,
  and Dean, J.
\newblock Outrageously large neural networks: The sparsely-gated
  mixture-of-experts layer.
\newblock In \emph{{ICLR} (Poster)}. OpenReview.net, 2017.

\bibitem[Shoeybi et~al.(2019)Shoeybi, Patwary, Puri, LeGresley, Casper, and
  Catanzaro]{shoeybi2019megatron}
Shoeybi, M., Patwary, M., Puri, R., LeGresley, P., Casper, J., and Catanzaro,
  B.
\newblock Megatron-lm: Training multi-billion parameter language models using
  model parallelism.
\newblock \emph{CoRR}, abs/1909.08053, 2019.

\bibitem[Stich(2019)]{stich2018localsgd}
Stich, S.~U.
\newblock Local {SGD} converges fast and communicates little.
\newblock In \emph{{ICLR} (Poster)}. OpenReview.net, 2019.

\bibitem[Su et~al.(2024)Su, Ahmed, Lu, Pan, Bo, and Liu]{Rope}
Su, J., Ahmed, M. H.~M., Lu, Y., Pan, S., Bo, W., and Liu, Y.
\newblock Roformer: Enhanced transformer with rotary position embedding.
\newblock \emph{Neurocomputing}, 568:\penalty0 127063, 2024.

\bibitem[Team et~al.(2025)Team, Bai, Bao, Chen, Chen, Chen, Chen, Chen, Chen,
  Chen, Chen, Cui, Ding, Dong, Du, Du, Du, Du, Fan, Feng, Fu, Gao, Gao, Gao,
  Gao, Gu, Guan, Guo, Guo, Hu, Hao, He, He, He, Hong, Hu, Hu, Huang, Huang,
  Huang, Jiang, Jiang, Jin, Kang, Lai, Li, Li, Li, Li, Li, Li, Li, Li, Li, Lin,
  Lin, Lin, Liu, Liu, Liu, Liu, Liu, Liu, Liu, Liu, Liu, Liu, Liu, Liu, Liu,
  Liu, Liu, Lu, Lu, Ma, Ma, Ma, Mao, Mei, Men, Miao, Pan, Peng, Qin, Qu, Shang,
  Shi, Shi, Song, Su, Su, Sun, Sung, Tang, Tao, Teng, Wang, Wang, Wang, Wang,
  Wang, Wang, Wang, Wang, Wang, Wang, Wang, Wang, Wang, Wang, Wang, Wang, Wang,
  Wei, Wei, Wu, Wu, Wu, Xiao, Xie, Xiong, Xu, Xu, Xu, Xu, Xu, Xu, Xu, Xu, Xu,
  Xu, Yan, Yan, Yang, Yang, Yang, Yang, Yang, Yao, Yao, Ye, Ye, Yin, Yu, Yuan,
  Yuan, Yuan, Zhan, Zhang, Zhang, Zhang, Zhang, Zhang, Zhang, Zhang, Zhang,
  Zhang, Zhang, Zhang, Zhao, Zhao, Zheng, Zheng, Zhou, Zhou, Zhou, Zhu, Zhuang,
  and Zu]{kimiteam2025kimik2openagentic}
Team, K., Bai, Y., Bao, Y., Chen, G., Chen, J., Chen, N., Chen, R., Chen, Y.,
  Chen, Y., Chen, Y., Chen, Z., Cui, J., Ding, H., Dong, M., Du, A., Du, C.,
  Du, D., Du, Y., Fan, Y., Feng, Y., Fu, K., Gao, B., Gao, H., Gao, P., Gao,
  T., Gu, X., Guan, L., Guo, H., Guo, J., Hu, H., Hao, X., He, T., He, W., He,
  W., Hong, C., Hu, Y., Hu, Z., Huang, W., Huang, Z., Huang, Z., Jiang, T.,
  Jiang, Z., Jin, X., Kang, Y., Lai, G., Li, C., Li, F., Li, H., Li, M., Li,
  W., Li, Y., Li, Y., Li, Z., Li, Z., Lin, H., Lin, X., Lin, Z., Liu, C., Liu,
  C., Liu, H., Liu, J., Liu, J., Liu, L., Liu, S., Liu, T.~Y., Liu, T., Liu,
  W., Liu, Y., Liu, Y., Liu, Y., Liu, Y., Liu, Z., Lu, E., Lu, L., Ma, S., Ma,
  X., Ma, Y., Mao, S., Mei, J., Men, X., Miao, Y., Pan, S., Peng, Y., Qin, R.,
  Qu, B., Shang, Z., Shi, L., Shi, S., Song, F., Su, J., Su, Z., Sun, X., Sung,
  F., Tang, H., Tao, J., Teng, Q., Wang, C., Wang, D., Wang, F., Wang, H.,
  Wang, J., Wang, J., Wang, J., Wang, S., Wang, S., Wang, Y., Wang, Y., Wang,
  Y., Wang, Y., Wang, Y., Wang, Z., Wang, Z., Wang, Z., Wei, C., Wei, Q., Wu,
  W., Wu, X., Wu, Y., Xiao, C., Xie, X., Xiong, W., Xu, B., Xu, J., Xu, J., Xu,
  L.~H., Xu, L., Xu, S., Xu, W., Xu, X., Xu, Y., Xu, Z., Yan, J., Yan, Y.,
  Yang, X., Yang, Y., Yang, Z., Yang, Z., Yang, Z., Yao, H., Yao, X., Ye, W.,
  Ye, Z., Yin, B., Yu, L., Yuan, E., Yuan, H., Yuan, M., Zhan, H., Zhang, D.,
  Zhang, H., Zhang, W., Zhang, X., Zhang, Y., Zhang, Y., Zhang, Y., Zhang, Y.,
  Zhang, Y., Zhang, Y., Zhang, Z., Zhao, H., Zhao, Y., Zheng, H., Zheng, S.,
  Zhou, J., Zhou, X., Zhou, Z., Zhu, Z., Zhuang, W., and Zu, X.
\newblock Kimi k2: Open agentic intelligence.
\newblock \emph{arXiv preprint arXiv:2507.20534}, 2025.

\bibitem[Tran et~al.(2025)Tran, Khiem, and Pham]{tran2025revisiting}
Tran, V.-T., Khiem, L.~H., and Pham, Q.-V.
\newblock Revisiting sparse mixture of experts for resource-adaptive federated
  fine-tuning foundation models.
\newblock In \emph{ICLR 2025 Workshop on Modularity for Collaborative,
  Decentralized, and Continual Deep Learning}, 2025.
\newblock URL \url{https://openreview.net/forum?id=IwNOUYgtuz}.

\bibitem[Vaswani et~al.(2017)Vaswani, Shazeer, Parmar, Uszkoreit, Jones, Gomez,
  Kaiser, and Polosukhin]{vaswani2017attention}
Vaswani, A., Shazeer, N., Parmar, N., Uszkoreit, J., Jones, L., Gomez, A.~N.,
  Kaiser, L., and Polosukhin, I.
\newblock Attention is all you need.
\newblock In \emph{{NIPS}}, pp.\  5998--6008, 2017.

\bibitem[Xie et~al.(2025)Xie, Luan, Cai, Yan, Chen, Xi, Fang, Shen, Wu, and
  Yuan]{xie2025dflmoe}
Xie, L., Luan, T., Cai, W., Yan, G., Chen, Z., Xi, N., Fang, Y., Shen, Q., Wu,
  Z., and Yuan, J.
\newblock dflmoe: Decentralized federated learning via mixture of experts for
  medical data analysis.
\newblock In \emph{Proceedings of the Computer Vision and Pattern Recognition
  Conference}, pp.\  10203--10213, 2025.

\bibitem[Yang \& Hu(2021)Yang and Hu]{yang2021tensorprogram}
Yang, G. and Hu, E.~J.
\newblock Tensor programs {IV:} feature learning in infinite-width neural
  networks.
\newblock In \emph{{ICML}}, volume 139 of \emph{Proceedings of Machine Learning
  Research}, pp.\  11727--11737. {PMLR}, 2021.

\bibitem[Zec et~al.(2020)Zec, Mogren, Martinsson, S{\"{u}}tfeld, and
  Gillblad]{zec2020federated}
Zec, E.~L., Mogren, O., Martinsson, J., S{\"{u}}tfeld, L.~R., and Gillblad, D.
\newblock Specialized federated learning using a mixture of experts.
\newblock \emph{CoRR}, abs/2010.02056, 2020.
\newblock \doi{10.48550/arXiv.2010.02056}.
\newblock URL \url{https://arxiv.org/abs/2010.02056}.

\bibitem[Zhai et~al.(2023)Zhai, He, Ma, Zong, Zhang, and
  Zhai]{zhai2023smartmoe}
Zhai, M., He, J., Ma, Z., Zong, Z., Zhang, R., and Zhai, J.
\newblock Smartmoe: Efficiently training sparsely-activated models through
  combining offline and online parallelization.
\newblock In \emph{{USENIX} {ATC}}, pp.\  961--975. {USENIX} Association, 2023.

\bibitem[Zhou et~al.(2022)Zhou, Lei, Liu, Du, Huang, Zhao, Dai, Chen, Le, and
  Laudon]{zhou2022expertchoice}
Zhou, Y., Lei, T., Liu, H., Du, N., Huang, Y., Zhao, V.~Y., Dai, A.~M., Chen,
  Z., Le, Q.~V., and Laudon, J.
\newblock Mixture-of-experts with expert choice routing.
\newblock In \emph{NeurIPS}, 2022.

\end{thebibliography}

\appendix
\phantomsection
\addcontentsline{toc}{section}{Appendix}
\section*{Appendix}

\section{Extended Related Works}\label{app:related}

\subsection{\moe architectures and scaling}
The sparsely-gated \moe layer \cite{shazeer2017moe} established the paradigm of conditional computation, utilizing trainable gating to activate a sparse subset of expert \ffns per token. In this work, we adopt the $\sigma$-\moe formulation \cite{csordas2023sigma}, which replaces the competitive softmax gating with independent sigmoid activations to better approximate dense feed-forward networks.

Formally, given an input token vector $\mathbf{x} \in \mathbb{R}^d$, the layer output $\mathbf{y}$ is the weighted sum of a sparse subset of expert networks $\{E_i\}_{i=1}^N$:
\begin{equation}
    \mathbf{y} = \sum_{i \in \mathcal{T}(\mathbf{x})} s_i(\mathbf{x}) E_i(\mathbf{x})\,,
\end{equation}
where $s_i(\mathbf{x}) = \sigma(\mathbf{w}_{g,i}^\top \mathbf{x})$ is the gating score for expert $i$, computed using the logistic sigmoid function $\sigma(\cdot)$ and a trainable gating vector $\mathbf{w}_{g,i}$. The set of active indices $\mathcal{T}(\mathbf{x})$ is determined by the Top-$K$ operator applied to the sigmoid scores:
\begin{equation}
    \mathcal{T}(\mathbf{x}) = \text{TopK}\left( \{\sigma(\mathbf{w}_{g,j}^\top \mathbf{x})\}_{j=1}^N, k \right)\,.
\end{equation}
Unlike standard Softmax gating, which introduces competition by normalizing scores to sum to one, the $\sigma$-MoE allows independent expert activations, mitigating instability.

This computation is \emph{conditional} because the experts $E_j(\mathbf{x})$ for indices $j \notin \mathcal{T}(\mathbf{x})$ are never executed. This mechanism decouples model capacity from compute cost: while the total parameter count scales linearly with the total number of experts $N$, the computational cost (FLOPs) per token scales only with the active set size $k$ (where $k \ll N$), outside of the router network and sigmoid operations. This enables massive parameter scaling with near-constant training/inference costs.

Subsequent systems scaled this approach: \texttt{GShard} introduced Top-2 gating with automatic sharding \cite{lepikhin2021gshard}, while \emph{Switch Transformers} demonstrated trillion-parameter training via Top-1 routing \cite{fedus2021switch}. \texttt{GLaM} further validated the compute efficiency of \moe architectures against dense baselines \cite{du2022glam}. Modern iterations, such as \texttt{DeepSeek-V2}, integrate \moe with optimized attention and KV-cache designs \cite{liu2024deepseekv2}. Theoretical advancements include fine-grained scaling laws \cite{krajewski2024fgmoe}, contextualized by foundational dense scaling laws \cite{kaplan2020scalinglaws, hoffmann2022chinchilla}. Beyond NLP, scalability has been confirmed in vision domains \cite{riquelme2021vmoe}.

\subsection{Routing and load balancing}
Load balancing is critical for maximizing \moe throughput and model quality. Approaches include \texttt{BASE}, which formulates assignment as a balanced linear optimization \cite{lewis2021base}; \texttt{Expert-Choice}, which inverts routing to allow experts to select tokens, ensuring perfect load balance \cite{zhou2022expertchoice}; and \texttt{Hash Layers}, which replace learned gating with fixed hashing functions \cite{roller2021hash}. In practice, training stacks employ temperature schedules, capacity factors, and auxiliary losses to prevent expert collapse and minimize token dropping \cite{fedus2021switch, lepikhin2021gshard}. Standard diagnostics—including routing entropy, load coefficients of variation, and overflow rates—are essential for monitoring these stability dynamics.

\subsection{Systems support for \moe execution}
Specialized frameworks optimize \moe execution through all-to-all communication, kernel fusion, and parallelization strategies. \texttt{DeepSpeed-MoE} and \texttt{Tutel} enable efficient expert parallelism and hierarchical communication \cite{rajbhandari2022deepspeedmoe, hwang2022tutel}, while \texttt{MegaBlocks} introduces dropless block-sparse kernels to mitigate padding inefficiencies \cite{gale2022megablocks}. \texttt{FasterMoE} and \texttt{SmartMoE} further optimize end-to-end training via dynamic hybrid parallelization schemes \cite{he2022fastermoe, zhai2023smartmoe}. Crucially, these systems are designed for high-bandwidth intra-\dc fabrics; our work adapts these principles to the bandwidth-constrained inter-\dc regime.

\subsection{Low-bandwidth distributed and federated pre-training}
To accommodate low-bandwidth environments, federated and \localsgd methods reduce synchronization frequency \cite{mcmahan2017fedavg, stich2018localsgd}, with extensions like \texttt{FedProx} and \fedopt addressing data heterogeneity \cite{li2020fedprox, reddi2020fedopt}.

\diloco applies these principles to \texttt{LM} pre-training, utilizing large local steps and outer momentum to train across weakly connected device islands \cite{douillard2023diloco}, a capability operationalized by \texttt{OpenDiLoCo} \cite{jaghouar2024opendiloco}. \photon further demonstrates federated pre-training up to 7B parameters with perplexity parity and significantly reduced communication \cite{sani2024photon}. Additional optimization strategies include decoupling embeddings via \dept \cite{iacob2024dept}, desynchronizing optimizer states via \desloc \cite{iacob2025desloc}, and co-designing modular paths via \texttt{DiPaCo} \cite{douillard2024dipaco}. While these systems motivate our design, they rely on full replicas.

\subsection{Expert placement and topology awareness}
While intra-datacenter routing is well-studied, expert placement across heterogeneous wide-area networks (\wan) remains underexplored. Existing systems model placement primarily for intra-cluster efficiency \cite{hwang2022tutel, he2022fastermoe, gale2022megablocks}. Our work investigates whether selective expert replication can minimize \wan traffic without degrading quality. We evaluate \emph{fixed}, \emph{random}, and \emph{affinity-based} placement policies, analyzing their impact on stability (routing entropy), specialization, and load balance \cite{lewis2021base, zhou2022expertchoice, fedus2021switch}.

\subsection{Federated MoE}
Early research in Federated \moe focused on personalization, combining global generalists with local specialists \cite{zec2020federated, guo2021pfl}, or employing client-specific gating (\texttt{FedMix}) to enhance generalization \cite{reisser2021fedmix}. Subsequent systems extended this to Transformers: \texttt{FedMoE} assigns personalized sub-\moes to clients \cite{mei2024fedmoe}, while \texttt{A3SMoE} optimizes expert utilization across heterogeneous devices \cite{tran2025revisiting}. Decentralized approaches like \texttt{dFLMoE}, Learning@home/\texttt{DMoE}, and \texttt{SWARM} explore expert fusion, volunteer computing, and heterogeneous weak-link model parallelism \cite{xie2025dflmoe, ryabinin2020towards, ryabinin2023swarm}. However, these studies primarily address personalization, volunteer-device execution, inference constraints, or model-parallel scheduling; they do not address the payload size limitations inherent to communication-efficient federated pre-training.

\subsection{Optimization, privacy, and governance}
Our approach relies on established analyses of adaptive federated optimizers (\texttt{FedAdam}, \texttt{FedYogi}) and \localsgd \cite{reddi2020fedopt, stich2018localsgd}. While privacy mechanisms like secure aggregation \cite{bonawitz2017secagg, abadi2016dpsgd, kairouz2021advances} can be integrated, they introduce trade-offs in quality and complexity. Additionally, routing and placement decisions can be logged to ensure auditability and fairness in non-IID settings.

\subsection{Synthesis}
Prior work establishes that: (i) \moe architectures achieve dense-model quality with reduced compute \cite{fedus2021switch, du2022glam, liu2024deepseekv2}; and (ii) low-communication training can approach centralized baselines under \wan constraints \cite{douillard2023diloco, sani2024photon}. However, current cross-silo systems remain limited by the requirement to transmit full dense models. Our work bridges this gap, introducing a federated \moe framework that targets \emph{bytes per sync} through optimized expert placement, thereby shifting the studied trade-offs among perplexity, communication, and wall-clock time.

\section{Model Architecture and Configuration}\label{app:model_architecture}

\begin{figure}[ht]
    \centering
    \includegraphics[width=\linewidth]{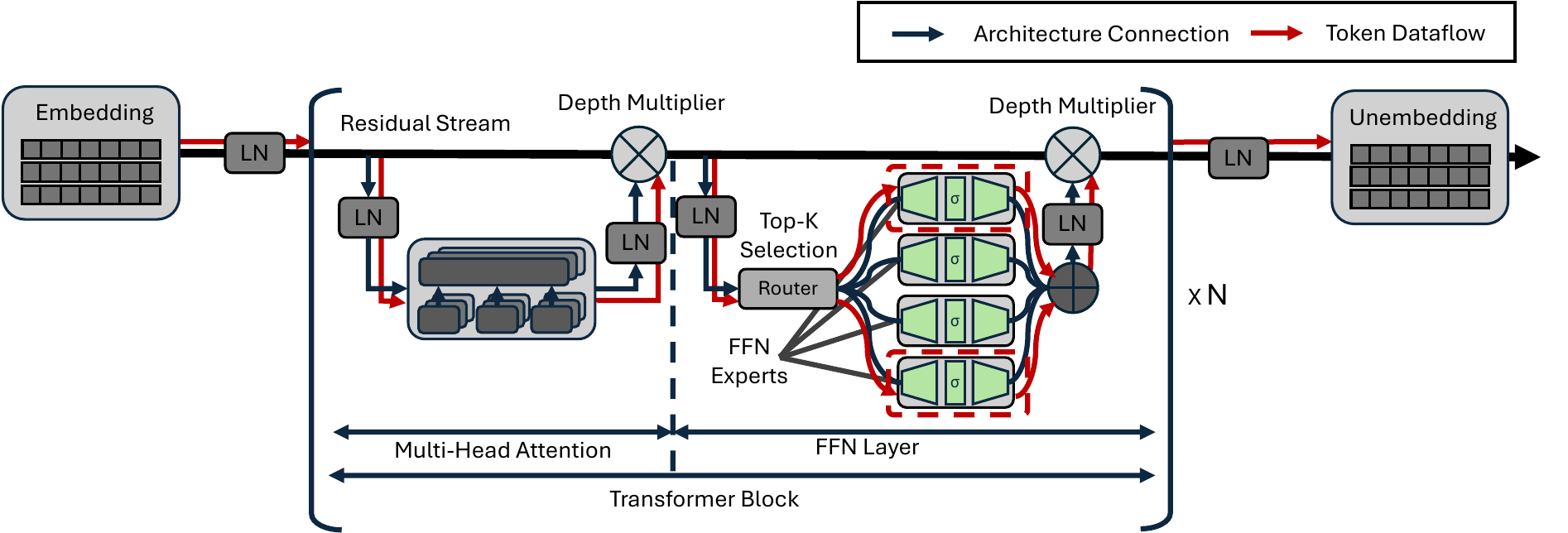}
    \caption{\textbf{\method Layer Architecture.} The diagram illustrates the Peripheral LayerNorm (Peri-LN) strategy, where normalization is applied exclusively to the branches feeding the Attention and MoE blocks (Pre-LN) and immediately following them (Post-LN). This design maintains a linear residual stream to preserve gradient magnitude. Additionally, the architecture includes a depth multiplier applied to the output of each computational block, as well as layer normalization on the output of the embedding layer and the input of the unembedding layer.}
    \label{fig:arch_diagram}
\end{figure}

To address the stability challenges of training large-scale models over bandwidth-constrained federated networks, \method adopts a specialized Transformer architecture. We depart from monolithic baselines by integrating \textbf{Sigma-MoE} ($\sigma$-MoE) routing with a \textbf{Peripheral LayerNorm} (Peri-LN) strategy. These choices maintain gradient flow and prevent router collapse in the \textit{Partial Replica} regime ($O_e < M$), where expert availability fluctuates due to random placement. The full architecture is shown in \cref{fig:arch_diagram}.

\subsection{Peripheral LayerNorm (Peri-LN)}
Standard Pre-LayerNorm (Pre-LN) allows unbounded growth in the residual stream, while Post-LayerNorm (Post-LN) impedes gradient propagation, causing instability in low-frequency synchronization settings.

We employ \textbf{Peripheral LayerNorm}~\cite{PeriLN}, which applies normalization exclusively to the residual \textit{branches}---both immediately preceding (Pre-LN) and following (Post-LN) the computational blocks---while preserving a linear, un-normalized main data path.

Formally, for a layer input $\mathbf{x}_l$ and computational block $\mathcal{F}$ (e.g., MHA or MoE):
\begin{align}
    \mathbf{x}_{l+1} = \mathbf{x}_l + \text{LN}(\mathcal{F}(\text{LN}(\mathbf{x}_l)))
\end{align}
This configuration ensures that inputs to the routing mechanism maintain unit variance, preventing dot-product logits from saturating the gating function, while maintaining a clear gradient highway from loss to input. We additionally apply a \textbf{Post-Embedding LayerNorm} to the initial token embeddings. As demonstrated in \cite{PeriLN}, this modification helps control the variance entering the first Transformer block, which has been shown to improve training stability. 

\subsection{Attention Mechanism}
We utilize Multi-Head Attention (MHA) enhanced with \textbf{Rotary Positional Embeddings (RoPE)}~\cite{Rope}. Positional information is injected into the Query ($\mathbf{Q}$) and Key ($\mathbf{K}$) vectors within the normalized branch (post-Peri-LN), ensuring relative positional information is preserved without dampening by residual stream statistics.

\subsection{\texorpdfstring{Federated Mixture-of-Experts ($\sigma$-MoE)}{Federated Mixture-of-Experts (sigma-MoE)}}
We replace the Feed-Forward Network (FFN) with a sparse Mixture-of-Experts layer based on the \textbf{$\sigma$-MoE} formulation.

\subsubsection{Expert Configuration and Fine-Grained Scaling}
To balance routing flexibility with memory fragmentation, we adopt a \textbf{Fine-Grained MoE} strategy~\cite{krajewski2024fgmoe}. We decompose the standard dense FFN ($d_{ff} = 4 d_{model}$) into $N_e$ experts with granularity $G=8$ and expansion $E=1$, leading to $N_{\text{e}} = G \cdot E$ experts. To maintain an iso-parameter count with the dense baseline, the expert dimension scales as:
\begin{equation}
    d_{expert} = \frac{4 d_{model}}{N_e}
\end{equation}
After each expert projection from $d_{model}$ to $d_{expert}$, the \silu nonlinearity is applied. For a $d_{model}=512$ configuration, we employ $N_e=16$ experts of dimension $d_{expert}=128$. This granular partitioning allows the router to approximate larger functional manifolds by combining specialized sub-experts, mitigating the impact of missing experts in partial replication.

\subsubsection{Sigmoid Routing}
Unlike competitive Softmax routers (e.g., Switch Transformer), \method employs \textbf{Sigmoid Routing} to enforce independent expert selection. This independence is critical for the \textit{Partial Replica} setting ($O_e < M$): the unavailability of a specific expert on a worker must not skew the probability distribution of the remaining available experts. The gating probability for expert $i$ is given by:
\begin{equation}
    p_i(\mathbf{x}) = \sigma(\mathbf{x}^\top \mathbf{w}_{r,i})
\end{equation}
After projecting scores from $d_{\text{model}}$ to $N_{\text{e}}$, the router selects the top-$k$ experts (typically $k=2$) based on these independent scores, with inputs normalized via the FFN's Peri-LN to ensure stable routing entropy.

\section{Hyperparameter Transferability and Implementation}\label{sec:systemdesign:transfer}

To avoid prohibitive hyperparameter retuning at scale, we leverage established transfer techniques. \emph{Maximal Update Parametrization} (\mup) enables the transfer of learning rates and schedules across model widths~\cite{yang2021tensorprogram}, while \completep extends this invariance to model depth via residual rescaling, preventing instability in deep Transformers~\cite{dey2025completep}. We note, however, that scaling the expert count $E$ requires specific adjustments to preserve feature learning efficiency~\cite{malasnicki2025muparam}.

\paragraph{Transfer protocols.}
We transfer the following hyperparameters from baseline configurations: global learning rate schedules (scaled via \mup/\completep), \adamw parameters ($\beta_1, \beta_2, \epsilon$), warmup duration, and router temperature. Conversely, regularization parameters (dropout, weight decay) and system-specific controls (overlap factor $\mathcal{O}_e$, reshuffling cadence $T$, local steps $K$) are treated as non-transferable and tuned individually at each scale. Consistent with \cite{malasnicki2025muparam}, we find that while \mup effectively handles width scaling, significantly increasing $E$ necessitates extending the learning rate warmup. Tuning was performed over a compute-optimal token budget~\cite{hoffmann2022chinchilla} using a linear warmup of $25\%$ without a subsequent cooldown phase. For \method, we evaluated learning rates $\eta \in \{0.001, 0.002, 0.004, 0.01, 0.02, 0.05\}$. Centralized baselines (dense and \moe) were tuned over $\eta \in \{0.01, 0.02, 0.03, 0.04, 0.07\}$. Across all architectures and aggregation methods, we observe that $\eta \approx 0.01$ yields the optimal learning rate, as shown in \cref{fig:completep_scaling}.

We use \completep to transfer the global learning rate schedule, \adamw settings ($\beta_1, \beta_2, \epsilon$), warmup duration, and router temperature across model scales for both our federated and centralized baseline experiments. This was repeated for different regularization factors and system-specific knobs, such as the overlapping factor $\mathcal{O}_e$, the shuffling cadence $T$, and the local steps $K$, as these do not scale with the \completep framework. In the sweeps, we found that the optimal learning rate across all our experiment types was $\eta\approx0.01$.

\begin{widefigure}[htb]
    \centering
    \includegraphics[width=1\textwidth, height=0.9\textheight, keepaspectratio]{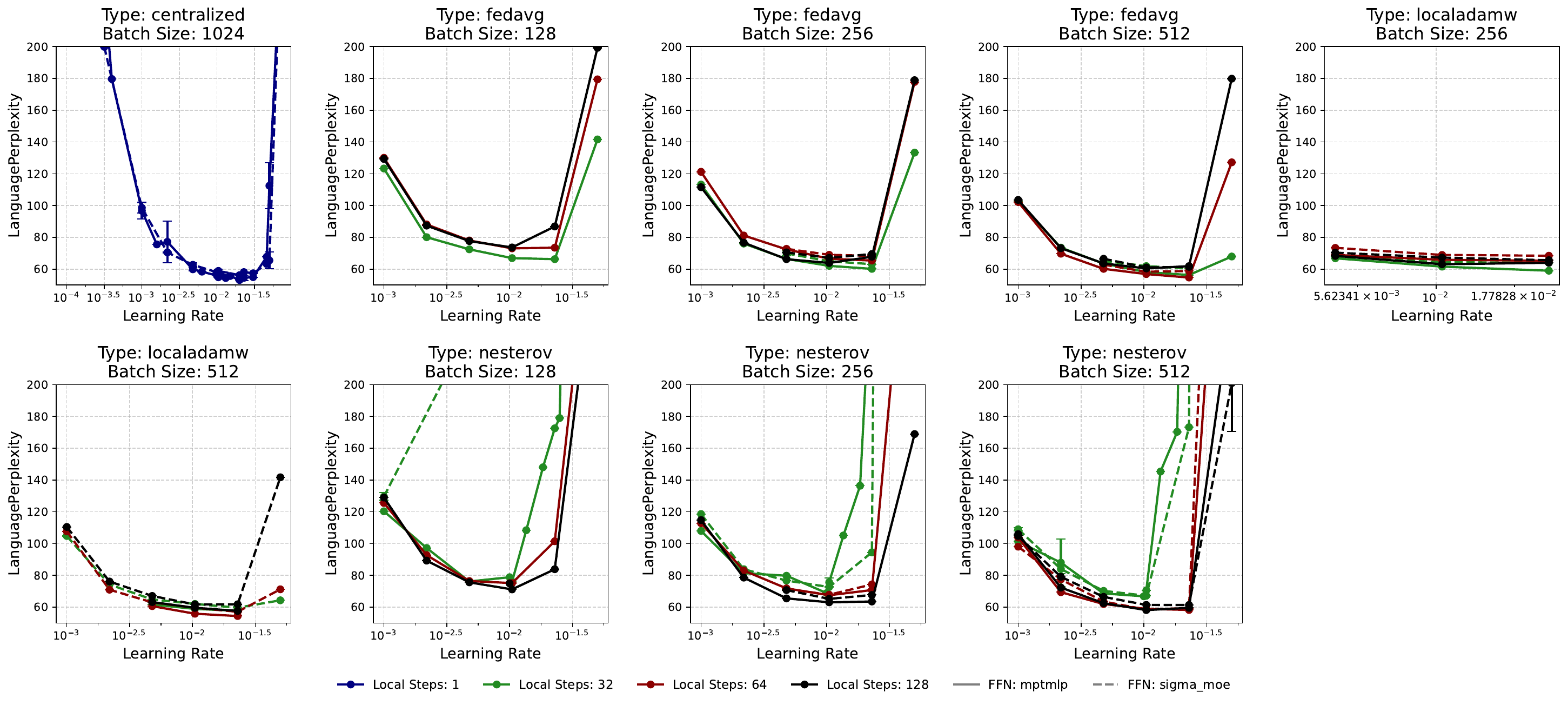}
    \caption{Learning rate tuning results for centralized (local step size of one) and federated baselines across varying local steps, architectures (dense vs. \moe), and aggregation methods. Across all configurations, a base learning rate of $\eta \approx 0.01$ consistently minimizes perplexity. We adopt this optimal rate for subsequent model scaling experiments.}
    \label{fig:completep_scaling}
\end{widefigure}

\paragraph{Implementation.}
We implemented \method using PyTorch's collective communication primitives. Standard frameworks such as \texttt{DistributedDataParallel} or \texttt{FSDP} are unsuitable for our setting, as they assume static, full-model replication and high-bandwidth intra-cluster interconnects. To support dynamic expert placement in \wan environments, we developed a custom coordination layer where the server explicitly manages the broadcast and aggregation of variable expert subsets at each synchronization round.

\section{Theoretical Foundations and Cost Modeling}\label{app:scalability_derivations}

This appendix serves as the formal theoretical underpinning for the \method system. We provide rigorous derivations for the cost models presented in \cref{sec:scalability-costs}, analyze the system's operational regimes, and formalize the trade-offs that govern the "Sweet Spot" for distributed \moe training.

\subsection{System Definitions and Parameters}

We model a synchronous training setup with $M$ homogeneous workers connected by a heterogeneous network topology. The system is defined by the global parameters listed in \cref{tab:system_params}.

\begin{table}[ht]
\centering
\caption{Global System and Model Parameters}
\label{tab:system_params}
\resizebox{0.95\hsize}{!}{
\begin{tabular}{l l l}
\toprule
\textbf{Symbol} & \textbf{Definition} & \textbf{Context} \\
\midrule
$M$ & Total number of workers (sites) & Infrastructure \\
$K$ & Synchronization interval (local steps) & Optimization \\
$B_{\text{eff}}$ & Effective WAN bandwidth & Infrastructure \\
$\mathcal{O}_e$ & Expert Overlap Factor ($\in [1, M]$) & \method Configuration \\
$N_{\text{le}}$ & Number of Local Experts per worker & Derived: $\frac{\mathcal{O}_e}{M} N_e$ \\
$\gamma$ & Reshuffling frequency ($1/T_{\text{shuffle}}$) & Random Placement \\
\bottomrule
\end{tabular}
}
\end{table}

\subsection{Operational Regimes and The Distributed Paradigm Shift}

To contextualize the \method architecture, we mathematically distinguish it from prior paradigms based on the constraints imposed on the synchronization interval $K$ and the expert overlap $\mathcal{O}_e$.

\paragraph{Regime 1: Distributed Data Parallelism (DDP).}
Defined by $\mathcal{O}_e = M$ (Full Replication) and $K=1$.
The communication payload is maximal ($P=0, R \approx 2$). Training is strictly bandwidth-bound on WANs:
\begin{equation}
    t_{\text{comm}}(\text{DDP}) \approx \frac{2 S_{\text{model}}}{B_{\text{eff}}} \gg t_{\text{comp}}
\end{equation}
Utilization approaches zero as $M$ increases across low-bandwidth links.

\paragraph{Regime 2: Federated / Low-Frequency (e.g., DiLoCo).}
Defined by $\mathcal{O}_e = M$ and $K \gg 1$.
Communication is amortized by $K$, but the payload remains the full model size $S_{\text{model}}$. The memory constraint is identical to DDP, limiting the maximum model size to the capacity of a single worker.

\paragraph{Regime 3: The \method Regime (Partial Replication).}
Defined by $\mathcal{O}_e < M$ (typically $\mathcal{O}_e \ll M$) and $K > 1$.
This regime introduces a "Communication-Sparsity" coupling. By partitioning experts, we reduce the steady-state memory and the synchronization payload linearly with $\mathcal{O}_e$. This breaks the "Full-Replica" barrier, allowing model size to scale linearly with $M$ while $t_{\text{comm}}$ decreases.

\subsection{Extended Computational Cost Modeling}\label{app:compute-derivation}

We refine the FLOPs derivation to explicitly account for both Router and FFN efficiency gains.

\paragraph{Router Efficiency.}
Standard gating networks project token embeddings of dimension $d_{\text{model}}$ to $N_e$ logits. In \method, tokens are only routed to locally resident experts. The router projection size is reduced from $N_e$ to $N_{\text{le}}$.
\begin{equation}
    F_{\text{gate}}(\mathcal{O}_e) = \frac{N_{\text{le}}}{N_e} F_{\text{gate}}^{\text{full}} = \frac{\mathcal{O}_e}{M} F_{\text{gate}}^{\text{full}}
\end{equation}
This yields a linear reduction in router FLOPs proportional to the partition fraction.

\paragraph{FFN Efficiency and Skip-Token Scaling.}
As derived in the Main Text, the FFN cost is scaled by $\sigma_{\text{st}}$, accounting for the "Skip-Token" mechanism when $N_{\text{le}} < k$.
\begin{equation}
    \sigma_{\text{st}} = \min\left(1, \frac{N_{\text{le}}}{N_e} \cdot \frac{N_e}{k}\right)
\end{equation}
Combining these, the total FLOPs per token is:
\begin{equation}
\resizebox{0.85\hsize}{!}{
$
    F_{\text{token}} \approx F_{\text{dense}} + \sum_{\ell=1}^L \left( \underbrace{\sigma_{\text{st}} \frac{\min(k, N_{\text{le}})}{k} F_{\text{FFN}}}_{\text{FFN Reduction}} + \underbrace{\frac{\mathcal{O}_e}{M} F_{\text{gate}}}_{\text{Router Reduction}} \right)$
}
\end{equation}

\subsection{Formal Derivation of Communication Factors}\label{app:comm-derivation}

We formally derive the probabilistic broadcast factor $P$ and the aggregation factor $R$ used in the Main Text.

\paragraph{The Aggregation Factor ($R$).}
For a specific expert replicated on $\mathcal{O}_e$ workers, Ring-AllReduce requires $2(\mathcal{O}_e - 1)$ transfers of size $S_{\text{expert}}/\mathcal{O}_e$.
Total data volume per expert: $\frac{2(\mathcal{O}_e - 1)}{\mathcal{O}_e} S_{\text{expert}}$.
This confirms the ring factor $R(\mathcal{O}_e) = \frac{2(\mathcal{O}_e - 1)}{\mathcal{O}_e}$ used in \cref{eq:experts_ring_factor}. Note that for disjoint experts ($\mathcal{O}_e=1$), $R=0$.

\paragraph{The Broadcast Factor ($P$).}
Under random placement with reshuffling frequency $\gamma$, we model the set difference between expert assignments $S_t$ and $S_{t+1}$.
Assuming uniform random assignment from $N_e$ experts:
\begin{compactitem}
    \item Probability of "Hit" (Expert $e \in S_{t+1}$ is in $S_t$):\\\quad $p_{\text{hit}} = \frac{|S_t|}{N_e} = \frac{\mathcal{O}_e}{M}$.
    \item Probability of "Miss" (Broadcast required):\\\quad $p_{\text{miss}} = 1 - \frac{\mathcal{O}_e}{M}$.
\end{compactitem}
The expected broadcast volume is proportional to the miss probability scaled by the reshuffling frequency $\gamma$:
\begin{equation}
    P(\mathcal{O}_e, M) = \gamma \left( 1 - \frac{\mathcal{O}_e}{M} \right)
\end{equation}
This derivation highlights the convexity of the cost function: $P$ maximizes when $\mathcal{O}_e$ is small (disjoint), which is exactly where $R$ is minimized. This necessitates $\gamma \ll 1$ (infrequent reshuffling) to make the disjoint regime viable.

\subsection{Optimization Surface and "Sweet Spot" Analysis}

We analyze the optimization surface of the Master Equation (\ref{eq:master_equation}) to identify the efficient "Sweet Spot" for training. We classify the operational space into four regions based on the variables $\mathcal{O}_e$ and $K$.

\begin{enumerate}
    \item \textbf{Region 1: Comm-Bound (High $\mathcal{O}_e$, Low $K$).}
    Approximating DDP. $t_{\text{comm}}$ dominates due to the full ring factor $R \approx 2$. Throughput is throttled by $B_{\text{eff}}$.

    \item \textbf{Region 2: Memory-Constrained (High $\mathcal{O}_e$, High $K$).}
    While communication is amortized by $K$, the high $\mathcal{O}_e$ prevents memory savings. Local batch size $B_l$ is constrained, preventing the "dilution" of communication costs (see Memory-Time Coupling in Main Text).

    \item \textbf{Region 3: The \method Sweet Spot (Low $\mathcal{O}_e$, High $K$).}
    Minimizing $\mathcal{O}_e$ drives $R \to 0$ and maximizes permissible $B_l$. This maximizes the Communication-to-Compute Ratio ($CCR$), enabling high throughput on low-bandwidth WANs. This is the optimal operating point.

    \item \textbf{Region 4: Placement Collapse (High $\gamma$, Low $\mathcal{O}_e$).}
    If reshuffling is too frequent ($\gamma \approx 1$), the Broadcast factor $P$ dominates $t_{\text{comm}}$, negating the benefits of disjoint aggregation. The system must operate with $\gamma \ll 1/t_{\text{comm}}$.
\end{enumerate}

\subsection{Model Falsifiability}

The theoretical models proposed here make specific, falsifiable predictions that can be validated empirically.

\paragraph{Prediction 1: Linearity of Broadcast Cost.}
We predict that for a fixed $M$, the broadcast volume during reshuffling will follow the curve $y = C \cdot (1 - \frac{\mathcal{O}_e}{M})$.

\paragraph{Prediction 2: Memory-Throughput Inverse Scaling.}
We predict that the maximum throughput (Tokens/Sec) scales super-linearly with $1/\mathcal{O}_e$ due to the compounding effects of reduced payload size and increased local batch size $B_l$.

\clearpage
{
\PaperIfTwoColumn{\onecolumn}{}
\section{Additional Results}

\begin{table}[H]
    \centering
        \caption{Evaluation for \method with \localadamw and \diloco across different configurations of workers $M$ and synchronization frequencies $K$.}
    \label{tab:exp0_extended_results}
    \resizebox{0.8\textwidth}{!}{            \begin{minipage}{0.48\linewidth}
        \centering
        \caption{\localadamw}
        \label{tab:sub_localadam}
        \resizebox{\linewidth}{!}{            \begin{tabular}{lccc}
                \toprule
                & \multicolumn{3}{c}{$K$} \\
                \cmidrule(lr){2-4}
                $M$ & 32 & 64 & 128 \\
                \midrule
                2 & $105.03 \pm 14.67$ & $104.70 \pm 14.80$ & $106.91 \pm 14.99$ \\
                4 & $103.91 \pm 14.43$ & $104.47 \pm 14.74$ & $111.89 \pm 15.57$ \\
                8 & $106.30 \pm 14.92$ & $112.99 \pm 15.71$ & $159.71 \pm 22.41$ \\
                \bottomrule
            \end{tabular}        }
    \end{minipage}
    \hfill
    \begin{minipage}{0.48\linewidth}
        \centering
        \caption{\diloco}
        \label{tab:sub_diloco}
                \resizebox{\linewidth}{!}{            \begin{tabular}{lccc}
                \toprule
                & \multicolumn{3}{c}{$k$} \\
                \cmidrule(lr){2-4}
                $M$ & 32 & 64 & 128 \\
                \midrule
                2 & $118.01 \pm 16.06$ & $117.94 \pm 15.85$ & $507.21 \pm 74.84$ \\
                4 & $137.27 \pm 18.35$ & $124.17 \pm 16.43$ & $420.49 \pm 65.81$ \\
                8 & $145.64 \pm 19.69$ & $133.37 \pm 17.77$ & $647.11 \pm 94.47$ \\
                \bottomrule
            \end{tabular}        }
    \end{minipage}}
\end{table}

\begin{table}[H]
\centering
\caption{Evaluation for \method with \localadamw and \diloco across different configurations of workers $M$, synchronization frequencies $K$, overlapping factors $O_e$ and placement strategies.}
\label{tab:exp1_results}

    \resizebox{0.8\textwidth}{!}{\begin{tabular}{lllccccc}
\toprule
 &  &  &  & \multicolumn{4}{c}{$O_e$} \\
\cmidrule(lr){5-8}
Strategy & $K$ & Placement Strategy &  $M$ & 1 & 2 & 4 & 8 \\
\midrule
\multirow[t]{18}{*}{\localadamw} & \multirow[t]{6}{*}{32} & \multirow[t]{3}{*}{Fixed} & 2 & $111.45 \pm 15.52$ & $105.03 \pm 14.67$  & - & - \\
 &  &  & 4 & $121.42 \pm 16.55$ & $112.00 \pm 15.45$ & $103.91 \pm 14.43$ & - \\
 &  &  & 8 & $135.74 \pm 18.57$ & $122.49 \pm 17.16$ & $114.69 \pm 16.08$ & $106.91 \pm 14.99$  \\
\cline{3-8}
 &  & \multirow[t]{3}{*}{Random} & 2 & $116.43 \pm 16.06$ & $105.03 \pm 14.67$ & - & - \\
 &  &  & 4 & $124.47 \pm 17.03$ & $113.65 \pm 15.81$ & $103.91 \pm 14.43$ & - \\
 &  &  & 8 & $140.37 \pm 19.07$ & $124.56 \pm 17.32$ & $115.48 \pm 15.91$ & $106.30 \pm 14.92$ \\
\cline{2-8} \cline{3-8}
 & \multirow[t]{6}{*}{64} & \multirow[t]{3}{*}{Fixed} & 2 & $112.38 \pm 15.53$ & $104.70 \pm 14.80$ & - & - \\
 &  &  & 4 & $130.79 \pm 17.67$ & $119.53 \pm 16.34$ & $104.47 \pm 14.74$ & - \\
 &  &  & 8 & $141.48 \pm 19.31$ & $132.39 \pm 18.30$ & $121.36 \pm 17.02$ & $112.99 \pm 15.71$ \\
\cline{3-8}
 &  & \multirow[t]{3}{*}{Random} & 2 & $119.73 \pm 16.79$ & $104.70 \pm 14.80$ & - & - \\
 &  &  & 4 & $129.74 \pm 17.72$ & $114.88 \pm 15.88$ & $104.47 \pm 14.74$ & - \\
 &  &  & 8 & $145.92 \pm 19.68$ & $131.02 \pm 18.31$ & $121.04 \pm 16.54$ & $112.99 \pm 15.71$ \\
\cline{2-8} \cline{3-8}
 & \multirow[t]{6}{*}{128} & \multirow[t]{3}{*}{Fixed} & 2 & $120.71 \pm 16.53$ & $106.91 \pm 14.99$ & - & - \\
 &  &  & 4 & $143.72 \pm 19.07$ & $123.50 \pm 16.91$ & $111.89 \pm 15.57$ & - \\
 &  &  & 8 & $181.07 \pm 23.70$ & $156.84 \pm 21.39$ & $138.88 \pm 19.34$ & $159.71 \pm 22.41$ \\
\cline{3-8}
 &  & \multirow[t]{3}{*}{Random} & 2 & $128.38 \pm 17.88$ & $106.91 \pm 14.99$ & - & - \\
 &  &  & 4 & $148.62 \pm 19.81$ & $125.36 \pm 17.15$ & $111.89 \pm 15.57$ & - \\
 &  &  & 8 & $195.09 \pm 26.19$ & $146.16 \pm 19.65$ & $131.86 \pm 18.16$ & $159.71 \pm 22.41$ \\
\cline{1-8} \cline{2-8} \cline{3-8}
\multirow[t]{18}{*}{\diloco} & \multirow[t]{6}{*}{32} & \multirow[t]{3}{*}{Fixed} & 2 & $131.69 \pm 17.99$ & $118.01 \pm 16.06$ & - & - \\
 &  &  & 4 & $142.94 \pm 18.82$ & $130.58 \pm 17.58$ & $137.27 \pm 18.35$ & - \\
 &  &  & 8 & $168.39 \pm 21.84$ & $149.95 \pm 20.03$ & $145.17 \pm 19.42$ & $145.64 \pm 19.69$ \\
\cline{3-8}
 &  & \multirow[t]{3}{*}{Random} & 2 & $142.38 \pm 18.67$ & $118.01 \pm 16.06$ & - & - \\
 &  &  & 4 & $157.49 \pm 20.78$ & $136.01 \pm 18.03$ & $137.27 \pm 18.35$ & - \\
 &  &  & 8 & $170.65 \pm 22.94$ & $180.43 \pm 24.33$ & $160.62 \pm 21.63$ & $145.64 \pm 19.69$ \\
\cline{2-8} \cline{3-8}
 & \multirow[t]{6}{*}{64} & \multirow[t]{3}{*}{Fixed} & 2 & $144.15 \pm 18.66$ & $117.94 \pm 15.85$ & - & - \\
 &  &  & 4 & $154.73 \pm 20.29$ & $142.25 \pm 18.41$ & $124.17 \pm 16.43$ & - \\
 &  &  & 8 & $187.66 \pm 23.94$ & $158.91 \pm 20.59$ & $149.30 \pm 19.85$ & $133.37 \pm 17.77$ \\
\cline{3-8}
 &  & \multirow[t]{3}{*}{Random} & 2 & $152.42 \pm 19.27$ & $117.94 \pm 15.85$ & - & - \\
 &  &  & 4 & $155.17 \pm 19.95$ & $136.46 \pm 17.82$ & $124.17 \pm 16.43$ & - \\
 &  &  & 8 & $182.99 \pm 23.34$ & $158.05 \pm 20.29$ & $144.43 \pm 19.23$ & $133.37 \pm 17.77$ \\
\cline{2-8} \cline{3-8}
 & \multirow[t]{6}{*}{128} & \multirow[t]{3}{*}{Fixed} & 2 & $289.86 \pm 41.29$ & $507.21 \pm 74.84$ & - & - \\
 &  &  & 4 & $342.25 \pm 47.39$ & $383.65 \pm 57.42$ & $420.49 \pm 65.81$ & - \\
 &  &  & 8 & $531.75 \pm 75.05$ & $384.71 \pm 54.90$ & $748.52 \pm 105.02$ & $647.11 \pm 94.47$ \\
\cline{3-8}
 &  & \multirow[t]{3}{*}{Random} & 2 & $403.32 \pm 59.70$ & $507.21 \pm 74.84$ & - & - \\
 &  &  & 4 & $282.21 \pm 40.20$ & $430.88 \pm 63.02$ & $420.49 \pm 65.81$ & - \\
 &  &  & 8 & $533.30 \pm 71.11$ & $314.18 \pm 44.39$ & $424.59 \pm 65.96$ & $647.11 \pm 94.47$ \\
\bottomrule
\end{tabular}
}
\end{table}
}
\PaperIfTwoColumn{\twocolumn}{}

\clearpage

\PaperIfTwoColumn{\onecolumn}{}
\section{Reproducibility Details}\label{app:reproducibility}

This appendix expands the compact experimental description in \cref{sec:experimental-setting}. The entries below report settings that are needed to reproduce the experiments.

\begin{widetable}[ht]
    \centering
    \scriptsize
    \setlength{\tabcolsep}{4pt}
    \caption{Reproducibility summary for the experimental setting. Values are grouped by the setting they reproduce.}
    \label{tab:reproducibility}
    \begin{tabular}{p{0.21\textwidth}p{0.73\textwidth}}
        \toprule
        \textbf{Item} & \textbf{Reproducibility detail} \\
        \midrule
        Model configurations & Five scales are used: Small Proxy (54M), Medium (150M), Large (2B), XL (13B), and XXL (100B). The larger configurations scale width and depth from the 54M proxy while keeping $d_{head}=64$, expansion factor $4$, expert hidden size $128$, and active expert ratio $0.25$. \\
        Tokenization and context & Text is tokenized with \texttt{openai/gpt-oss-120b}; the tokenizer vocabulary size is $200{,}019$. All reported training configurations use sequence length $2048$. \\
        Training corpus & The pre-training mixture uses \texttt{HuggingFaceTB/smollm-corpus} subsets \texttt{fineweb-edu-dedup} and \texttt{python-edu} in a $90/10$ stream proportion. \\
        Evaluation data & Evaluation uses the C4 \texttt{en} validation split. The reported standard deviations are computed across runs with different seeds. \\
        Batch and budget & Global batch size is $1024$ sequences. For $M\in\{2,4,8\}$ workers, local batch size is $1024/M$. The proxy training budget targets $20$ tokens per parameter; after synchronization-boundary rounding, this corresponds to $640$ optimizer batches, $160$ warmup batches, and approximately $1.34$B realized tokens for the 54M proxy. \\
        Optimization & Local training uses \adamw with weight decay $0$, betas $(0.9,0.95)$, and $\epsilon=10^{-8}$. The proxy learning-rate sweep covers $\{0.01,0.01144714,0.01310371,0.015\}$ before selecting transferred rates for the main sweeps. \diloco/Nesterov uses server learning rate $0.7$ and momentum $0.9$; \localadamw synchronizes first and second optimizer moments. \\
        Sweep grid & The main grid uses $K\in\{32,64,128\}$ local steps, $M\in\{2,4,8\}$ workers, seeds $\{42,123,456\}$, and full participation. Full-replica \moe baselines use $\mathcal{O}_e=M$; fixed and random placement sweeps use the valid lower powers of two for each $M$; skip-token sweeps additionally include full overlap and set \texttt{skip\_tokens=True}. \\
        Hardware reporting & Experiments were run on NVIDIA GPU cluster nodes. \\
        WAN parameters & The system analysis uses controlled \wan parameters of \SI{1}{\giga\bit\per\second} and \SI{10}{\giga\bit\per\second} with \SI{50}{\milli\second} latency, no time-varying jitter, no asymmetric bandwidth, and no persistent stragglers. \\
        \bottomrule
    \end{tabular}
\end{widetable}

\PaperIfTwoColumn{\twocolumn}{}
\clearpage

\PaperIfTwoColumn{\onecolumn}{}
\section{Training Algorithm}

    \captionof{algorithm}{\method - Simplified View}
    \label{alg:simplified_fomoe}
        \hrule height 0.4pt
    \vspace{0.5em}

    \begin{algorithmic}[1]
    \footnotesize
    \Require \textbf{Model Partition and Groups}
        \State $\theta = \{\theta_{\text{dense}}, \theta_{\text{sparse}}\}$ — params partitioned into dense and sparse components
        \State $\mathcal{W}_{\text{all}} = \{1, \dots, M\}$ — set of all workers
    \Require \textbf{Hyper-parameters}
        \State $\theta_0$— initial params
        \State $\beta_1, \beta_2 \in [0,1)$ — decay rates AdamW momenta
        \State $\texttt{LocalOptimization}$ — update model parameters locally with AdamW and skip tokens
        \State $\texttt{OuterOpt}$ — update aggregated params using an outer optimizer (e.g., Adam)
    \Ensure $\theta^T,\;u^{T-1},\;v^{T-1}$
        
    \State \textbf{for each worker} $m \in \mathcal{W}_{\text{all}}$: initialize $\theta^0, \{(u_m^{-1}, v^{-1}_m)\}^M_{m=1}$ \hfill\textcolor{gray}{\scriptsize initialize random initial model and all local momenta.}
    
    \For{$t = 0,\dots,T-1$}
        \Statex \textcolor{gray}{\scriptsize $T$ synchronization rounds}
        \State \textcolor{gray}{\textit{\scriptsize \quad 1. Global Calculation (happens once)}}
        \State $\mathcal{S} \gets \texttt{PlacementStrategy}(\theta_{\text{sparse}}^t, O_e, \text{Placement})$ \hfill\textcolor{gray}{\scriptsize generate allocation set $\mathcal{S} = \{s_0, \dots, s_{M-1}\}$}
        
        \State \textcolor{gray}{\textit{\scriptsize \quad 2. Vectorized / Simultaneous Assignment}}
        \State $\theta_{m, \text{sparse}}^t \gets s_m, \quad \forall s_m \in \mathcal{S}$ \hfill\textcolor{gray}{\scriptsize simultaneous distribution of sparse subsets}
        \State $\theta_{m, \text{dense}}^t \gets \theta_{\text{dense}}^t, \quad \forall m$ \hfill\textcolor{gray}{\scriptsize broadcast shared dense parameters}
        
        \State \textcolor{gray}{\textit{\scriptsize \quad 3. Model Compilation}}
        \State $\theta_m^t \gets \{ \theta_{m, \text{dense}}^t, \theta_{m, \text{sparse}}^t \}$ \hfill\textcolor{gray}{\scriptsize broadcast specialized models to workers}
        
        \ForAll{workers $m=0,\dots,M-1$ \textbf{in parallel}}
            \For{$k = 0, \dots K-1 $}
                \Statex \textcolor{gray}{\scriptsize $K$ local training steps}
                \State \texttt{LocalOptimization}$({\theta_m^t})$ \hfill\textcolor{gray}{\scriptsize AdamW optimization with skip tokens implementation}
            \EndFor
            \State $\Delta^{t+1}\gets$\texttt{Ring All-Reduce$({\{\theta_m^t}\}_{m=1}^M)$} \hfill\textcolor{gray}{\scriptsize aggregate dense parameters globally and sparse parameters within expert replica groups}
            \State $\theta^{t+1}\gets$\texttt{OuterOpt}$(\Delta^{t+1})$ \hfill\textcolor{gray}{\scriptsize perform outer optimization procedure}
        \EndFor
    \EndFor
    \end{algorithmic}
    
    \vspace{0.5em}
    \hrule height 0.4pt
    \vspace{0.5em}

The execution protocol proceeds in three phases per synchronization round ($T$). First, the \texttt{PlacementStrategy} generates a global allocation set $\mathcal{S}$, mapping expert subsets to workers according to the overlap factor $\mathcal{O}_e$; this step supports both static partitioning and dynamic reshuffling. Second, the system executes a vectorized assignment where dense parameters are broadcast globally, while sparse experts are routed exclusively to their assigned hosts. Finally, workers perform $K$ steps of \texttt{LocalOptimization} in parallel. Crucially, this phase utilizes a `skip-token' mechanism to bypass computations for non-resident experts, ensuring the computational savings of partial replication are realized. The round concludes with sparse aggregation, where gradients are synchronized only among the $\mathcal{O}_e$ replicas via \texttt{Ring All-Reduce} before the global state is updated.

\end{document}